\documentclass[final]{cvpr}

\usepackage{times}
\usepackage{epsfig}
\usepackage{graphicx}
\usepackage{amsmath}
\usepackage{amssymb}
\usepackage{float}
\usepackage{multirow}
\usepackage{enumitem}
\usepackage{graphicx}
\usepackage{booktabs}
\usepackage{multirow}
\usepackage{algorithm}
\usepackage{algpseudocode}

\usepackage[pagebackref=false,breaklinks=true,letterpaper=true,colorlinks,urlcolor=blue,citecolor=blue,linkcolor=blue,bookmarks=false]{hyperref}

\begin{document}
	
	\title{MetaCloth: Learning Unseen Tasks of Dense Fashion Landmark Detection\\ from a Few Samples}
	
	\author{Yuying Ge$^1$ \quad
		Ruimao Zhang$^2$ \quad
		Ping Luo$^1$\\
		{$^1$The University of Hong Kong} \quad \\
		{$^2$The Chinese University of Hong Kong (Shenzhen)}  \\
		\tt\small yuyingge@hku.hk \quad pluo@cs.hku.hk \quad
		\tt\small{ruimao.zhang@ieee.org}
	}
	\maketitle
	
	\begin{abstract}
	Recent advanced methods for fashion landmark detection are mainly driven by training convolutional neural networks on large-scale fashion datasets, which has a large number of annotated landmarks. However, such large-scale annotations are difficult and expensive to obtain in real-world applications, thus models that can generalize well from a small amount of labelled data are desired. We investigate this problem of few-shot fashion landmark detection, where only a few labelled samples are available for an unseen task. This work proposes a novel framework named MetaCloth via meta-learning, which is able to learn unseen tasks of dense fashion landmark detection with only a few annotated samples. Unlike previous meta-learning work that focus on solving ``$N$-way $K$-shot'' tasks, where each task predicts $N$ number of classes by training with $K$ annotated samples for each class ($N$ is fixed for all seen and unseen tasks), a task in MetaCloth detects $N$ different landmarks for different clothing categories using $K$ samples, where $N$ varies across tasks, because different clothing categories usually have various number of landmarks. Therefore, numbers of parameters are various for different seen and unseen tasks in MetaCloth. MetaCloth is carefully designed to dynamically generate different numbers of parameters for different tasks, and learn a generalizable feature extraction network from a few annotated samples with a set of good initialization parameters. Extensive experiments show that MetaCloth outperforms its counterparts by a large margin.
\end{abstract}

\begin{figure}[t]
	\begin{center}
		\includegraphics[width=1\linewidth]{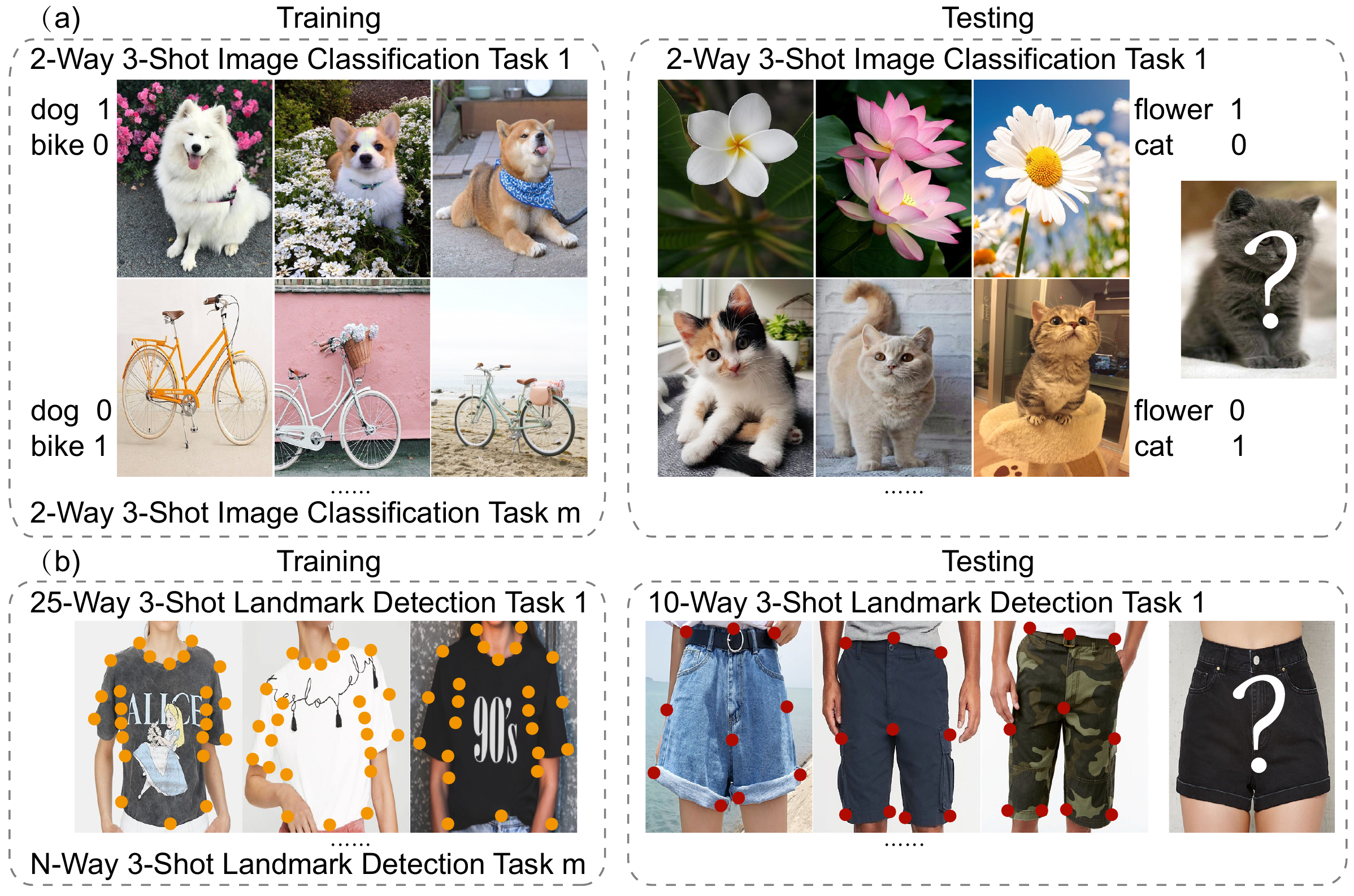}
	\end{center}
	\vspace{-10pt}
	\caption{Example of (a) meta-learning for image classification, where both training and testing consist of 2-way 3-shot tasks. During test, it aims to classify 2 unseen classes with 3 samples for each class in a task; (b) meta-learning for fashion landmark detection, where training and testing consist of $N$-way 3 shot tasks and $N$ varies across tasks. For example,  a 25-way 3-shot task in training learns to detect 25 landmarks with 3 samples for ``short sleeve shirt'' while a 10-way 3-shot task in the test aims to detect 10 landmarks with 3 samples for ``shorts''. After training a model on landmark annotations of \eg upper-body clothes, it can adapt to detect landmarks of lower-body clothes with only a few samples.}
	\vspace{-10pt}
	\label{fig:Overview}
\end{figure}

\section{Introduction}

\renewcommand\arraystretch{1.1}
\newcommand{\tabincell}[2]{\begin{tabular}{@{}#1@{}}#2\end{tabular}}  
\begin{table*}\small\centering
	\vspace{-10pt}
	\caption{Comparisons of the setups in different few-shot applications that use meta-learning, including (a) image classification, (b) object detection, (c) semantic segmentation, and (d) dense fashion landmark detection (ours). We see that prior arts (a-c) employed a ``$N$-way $K$-shot'' setup by producing predictions of $N$ classes trained on $K$ samples for each class. ${S}$ is a support set for each task, containing a subset of samples and their labels sampled from a dataset $\mathcal{D}$. In (d), a ``$N_c$-way $K$-shot'' setup is adopted for dense fashion landmark detection where each task learns to detect $N_c$ landmarks and $N_c$ varies in different tasks.} 
	\scalebox{0.8}{
		\begin{tabular}{c|c|c|c|c}
			\hline
			\multicolumn{1}{c|}{Few-shot Applications}&\multicolumn{1}{c|}{Image Classification }&\multicolumn{1}{c|}{Object Detection}&\multicolumn{1}{c|}{Semantic Segmentation}&\multicolumn{1}{c}{Dense Fashion Landmark Detection}\\
			\hline
			\multirow{1}*{Task ${T}$}&$N$-way $K$-shot ~\cite{sun2019meta}&$N$-way $K$-shot ~\cite{wu2020meta}& $N$-way $K$-shot ~\cite{wang2019panet}&$N_c$-way $K$-shot\\
			\hline
			\multirow{1}*{Support Set ${S}$}
			&\tabincell{c}{${S}=\{(I_k^n,y_k^n)\}_{n=1,k=1}^{N,K}$;\\where $I_k^n$: $k^{\mathrm{th}}$ image of the \\$n^{\mathrm{th}}$ class.\\$y_k^n$: label of $I_k^n$;}
			&\tabincell{c}{${S}=\{(o_k^n,y_k^n)\}_{n=1,k=1}^{N,K}$;\\where $o_k^n$: $k^{\mathrm{th}}$ object of the \\$n^{\mathrm{th}}$ class;\\$y_k^n$: label of $o_k^n$;}
			&\tabincell{c}{${S}=\{(s_k^n,y_k^n)\}_{n=1,k=1}^{N,K}$;\\where $s_k^n$: $k^{\mathrm{th}}$ segmentation of the \\$n^{\mathrm{th}}$ class;\\$y_n^k$: label of $s_k^n$;}
			&\tabincell{c}{${S}=\{(I_k^c,Y_k^c)\}_{k=1}^{K}$;\\where $I_k^c$: $k^{\mathrm{th}}$ image of the $c^{\mathrm{th}}$ category \\and $I_k^c$ has $N_c$ landmarks;\\$Y_k^c$: a set of $N_c$ labels and each label\\ is the coordinate of a landmark;}
			\\
			\hline
	\end{tabular}}
	\vspace{-10pt}
	\label{tab:formulation}
\end{table*}

Detecting dense clothing landmarks is important for fashion image understanding, because the keypoints on the clothes provide discriminative features that enable various applications of fashion image analysis such as clothes recognition, retrieval, and virtual try-on~\cite{liu2016deepfashion,ge2019deepfashion2,tryon, ge2021parser, ge2021disentangled}. 
However, recent success of dense fashion landmark detection is driven by training convolutional neural networks (CNNs) on large-scale datasets such as DeepFashion2~\cite{ge2019deepfashion2}, which has a large number of annotated landmarks. 
The training of the above CNNs requires large amounts of annotated data to minimize a loss function $\ell$ with respect to a set of network parameters $\theta$, ${{\min}_\theta~}\mathbb{E}_{(I_i,y_i)\sim{\mathcal{D}}}[{\ell}_\theta(I_i,y_i)]$, where $(I_i,y_i)$ denotes an image and its label sampled from a large-scale dataset $\mathcal{D}$ and $\mathbb{E}[\cdot]$ represents expectation.
However, such annotations demand considerable labor, as labelling dense landmark locations of clothing images are critically time-consuming and expensive in practice.
In order to get rid of the dependence on a large number of annotated clothing landmarks, a fashion landmark detection model that can generalize well from a small amount of labelled data is desired.
We investigate this problem of few-shot fashion landmark detection, where only a few annotated samples are available for an unseen task (\eg an unseen clothing category).

A straightforward way to solve the above problem is to apply meta-learning, which trains a model on a variety of learning tasks, such that it can solve new few-shot tasks using only a small number of annotated samples.
Specifically, meta-learning aims to minimize the loss function with respect to a set of tasks $\mathcal{T}=\{T_i\}$, ${{\min}_\theta~}\mathbb{E}_{{T}\sim{p({T})}}[{\ell}_\theta({T})]$, where each task ${T_i}$ is trained on a subset of images and their labels. And ${T_i}$ is sampled from  a distribution of tasks $p({T})$.
In this case, the learned parameters $\theta$ would provide intermediate representation that is more easily to generalize to unseen tasks than the conventional supervised learning.
The key idea is to leverage a large number of few-shot tasks to learn how to adapt a model to a new task where only a few labelled samples are available.
We summarize the existing meta-learning applications in Table~\ref{tab:formulation}, including few-shot image classification~\cite{finn2017model,nichol2018first,gidaris2019generating,sun2019meta,baik2020meta}, few-shot object detection~\cite{kang2019few,wang2019meta,fu2019meta,wu2020meta}, and few-shot semantic segmentation~\cite{dong2018few,wang2019panet,baik2020meta,zhang2020sg,pambala2020sml}. We observe that they focused on training many tasks in a ``$N$-way $K$-shot'' setup, where each task predicts $N$ classes by training on $K$ images for each class. Here the number of classes $N$ is fixed as the same value for all seen and unseen tasks.
For example, in meta-learning for few-shot image classification, a task is trained by sampling a small support dataset ${S}$ from the training set $\mathcal{D}$. ${S}$ contains a subset of $K$ images and their labels, ${S}=\{(I_k^n,y_k^n)\}_{n=1,k=1}^{N,K}$, rendering a ``$N$-way $K$-shot'' setup.
Every task predicts $N$ number of classes, where the value of $N$ is fixed for all seen and unseen classifiers as shown in Fig.~\ref{fig:Overview}.
Another examples are meta-learning for few-shot object detection and few-shot semantic segmentation, where each object bounding box or segmentation mask belongs to one of the $N$ object classes and the value $N$ is also fixed for all object detectors or mask predictors.

In this paper, we tackle a more challenging problem, ``$N_c$-way $K$-shot'' fashion landmark detection, which aims to detect $N_c$ fashion landmarks using $K$ samples. The support set is written as ${S}=\{(I_k^c,Y_k^c)\}_{k=1}^{K}$, where $I_k^c$ denotes the $k^{th}$ image of the $c^{th}$ clothing category and it has $N_c$ landmarks, while $Y_k^c$ indicates a set of coordinates of the landmarks in this image.
Unlike the dominant setup in existing meta-learning applications that the numbers of predictions $N$ (\eg \#classes) are the same for all tasks, different seen and unseen detectors predict different $N_c$ numbers of landmark locations as shown in Fig.~\ref{fig:Overview}, because various clothing categories have various number of landmark definitions (\eg $N_c=25$ for ``short  sleeve  shirt'' and $N_c=10$ for ``shorts'' in the DeepFashion2 dataset).
As $N_c$ varies across seen and unseen tasks, they have different numbers of parameters, such that the parameter space is changed from time to time in different tasks. This makes few-shot fashion landmark detection model more difficult to learn intermediate representation that can efficiently adapt to new tasks.
While \cite{metadataset} introduced a similar setup with ``$N_c$-way $K$-shot tasks'', where $N_c$ varies, it only evaluated popular baselines for few-shot image classification with minor modifications. For example, to handle image classification with variable ways, it simply initialized the classification weights and bias of MAML~\cite{finn2017model} to zero instead of learning the initialization, leading to sub-optimal performance. An elaborately designed and effective method to solve the challenging few-shot fashion landmark detection is still to be explored.

To address the above difficulty, we propose a new meta-learning framework for few-shot dense fashion landmark detection, named MetaCloth, enabling learning unseen tasks of fashion landmark detection from just a few samples.
For example, a convolutional neural network (CNN) produced by MetaCloth can be trained on landmark annotations of upper-body clothes (\eg ``short sleeve shirt''), and then effectively adapts to detect landmarks of lower-body clothes (\eg ``shorts'') with merely a few annotations.
This implies that MetaCloth enables a learned model generalized to arbitrary clothing category with any number of landmarks by using only a few annotated images.
As shown in Table~\ref{tab:formulation}, this ``$N_c$-way $K$-shot'' setup defines a dynamic parameter space, where the number of parameters will be changed for different tasks.
Therefore, landmark detectors for human body with fixed number of parameters are incapable in this setup.
Instead of freezing the parameter space, MetaCloth dynamically generates parameters for different number of landmark detectors when different tasks are presented.
After the parameters of the landmark detectors are predicted, it learns a generalizable feature extraction network from a few samples with a set of good initialization parameters, which are obtained through training a variety of tasks as prior knowledge.

Specifically, MetaCloth consists of a feature extraction network (FENet), landmark detectors (LD) and a parameter prediction network (PPNet) . 
FENet and LD constitute a complete fashion landmark detection model, where FENet extracts cloth-level features and LD perform pixel-wise classification on the cloth-level features and predict the location of fashion landmarks. 
Instead of fixing the architecture of landmark detectors, the parameters of LD are predicted by PPNet adaptively across tasks, to enable LD to predict different number of landmarks for different tasks with a dynamic parameter space.
For example, given a task with $N_c$ fashion landmarks, PPNet predicts parameters of $N_c$ landmark detectors taking $N_c$ landmark-level features as inputs, thus the architecture of landmark detectors is changed to make $N_c$ predictions.
During training, after the parameters of LD are predicted, the parameters of FENet are explicitly optimized for initialization, such that the feature extraction network can be effectively adapted to a new task.

This work has three main \textbf{contributions}.
\begin{itemize}[]
	\item We propose a novel few-shot dense fashion landmark detection method with meta-learning, named MetaCloth, which can learn to predict landmarks for unseen tasks from only a few annotated images. 
	To our knowledge, this is \textbf{the first work} for dense fashion landmark detection in a ``$N_c$-way $K$-shot'' setup.
	Different from the dominant setup in existing meta-learning applications in Table~\ref{tab:formulation}, where the number of classes $N$ is fixed for all seen and unseen tasks, our setup with varying $N_c$ classes is more realistic and challenging.
	\item An effective algorithm can dynamically generate parameters for different number of landmark detectors of unseen tasks and learn a highly generalizable feature extraction network with meta-learned initialization, to adapt to unseen scenarios from only a few samples (\eg from upper-body clothes to lower-body clothes).
	\item We carefully evaluate MetaCloth by designing many new benchmarks in DeepFashion2, which contains 13 different definitions of dense landmarks for 13 clothing categories.
	We construct eight competitive baselines through tailoring the existing few-shot learning methods.
	Extensive experiments demonstrate that MetaCloth outperforms its counterparts by a large margin.
\end{itemize}



\section{Related Work}
\subsection{Fashion Landmark Detection}
Different from human landmark detection~\cite{toshev2014deeppose,cao2019openpose}, which has a single category (human body), fashion landmark detection is to localize various landmarks for various clothing categories (\eg `shirt', `skirt'). 
Different categories have different landmark definitions. For example, DeepFashion2~\cite{ge2019deepfashion2} defined 25 landmarks for ``short sleeve shirt'' and 10 landmarks for ``shorts''.
Recent advances~\cite{liu2016deepfashion,wang2018attentive,li2019spatial,yu2019layout} in fashion landmark detection relied heavily on a huge amount of training data with manually annotated landmarks.
For example, DeepFashion~\cite{liu2016deepfashion} first introduced this task, where an alignment network leveraged pseudo-labels and auto-routing mechanism to extract features for different landmarks.
Non-local module was added in ~\cite{li2019spatial} to capture global dependency and utilize spatial information of landmarks.
Furthermore ~\cite{liu2016deepfashion,li2019spatial,yu2019layout,wang2018attentive} leveraged human knowledge among landmarks to enhance feature representation.
However, previous work explored fashion landmark detection in the regime of ordinary supervised learning, which required a large set of annotated data.
In contrast, MetaCloth investigates few-shot fashion landmark detection with meta-learning for the first time, where only a few annotated samples are provided for an unseen clothing category.

\subsection{ Few-shot Learning}
Few-shot learning aims to generalize to new tasks, which only contain a few samples.
Previous approaches utilized prior knowledge to augment data, such that the supervised information was enriched.
For example, \cite{kwitt2016one,hariharan2017low,liu2018feature,schwartz2018delta} transformed the training samples into several samples with variations. 
\cite{pfister2014domain,wu2018exploit,douze2018low} augmented data by selecting samples with the target label from a large data set, which was weakly labeled or unlabeled.

Besides data augmentation, lots of methods solved few-shot learning from the model perspective, which can be classified into four types:
a). Multitask Learning~\cite{yan2015multi,luo2017label,motiian2017few,benaim2018one}, which learned multiple  related tasks simultaneously by exploiting both task-generic and task-specific information.
b). Embedding Learning,~\cite{fink2004object,koch2015siamese,triantafillou2017few,oreshkin2018tadam} which embedded each sample to a lower-dimensional such that similar samples are close together while dissimilar samples can be more easily differentiated.
c). Learning with External Memory~\cite{xu2017few,zhu2018compound,cai2018memory,ramalho2019adaptive}, which extracted knowledge from the training set and stored it in an external memory. 
d). Generative Modelling~\cite{salakhutdinov2012one,rezende2016one,reed2017few,zhang2018metagan},  which estimated the probability distribution from the observed training samples.

An increasingly popular solution for few-shot learning is meta-learning, which leverages a large number of few-shot tasks in order to learn how to adapt a model to a new task.
In this work, we adopt meta-learning to solve the few-shot fashion landmark detection, which learns unseen tasks from a few samples after training a model on a variety of tasks.

\subsection{Meta Learning}
The goal of meta-learning is to train a model on a variety of learning tasks, such that it can solve new learning tasks using only a small number of training samples. 
Recently, meta-learning has gained increasing attention due to its superior performance in solving few-shot problems such as  image classification~\cite{finn2017model,nichol2018first,gidaris2019generating,sun2019meta,baik2020meta}, object detection~\cite{kang2019few,wang2019meta,fu2019meta,wu2020meta}, semantic segmentation~\cite{shaban2017one,dong2018few,wang2019panet,zhang2020sg,pambala2020sml}.
%
Existing meta-learning methods can be generally divided into three categories, including optimization-based, metric-based and parameter prediction methods. 

%
In the first category, the optimization-based methods~\cite{finn2017model,li2017meta,nichol2018first,rajeswaran2019meta,baik2020meta,fu2019meta,banerjee2020meta} learned a a good initialization so that a few gradient updates on its parameters would lead to good performance on unseen tasks.
For example, MAML~\cite{finn2017model} optimized parameters for initialization by differentiating through the inner loop optimization and Meta-SGD~\cite{li2017meta} further learned update direction and learning rate besides initialization.
However, these methods demanded fixed network architecture for each task, different from our setup where the number of parameters varies across tasks, thus are not flexible enough to solve few-shot fashion landmark detection.

In the second category, the metric-based methods ~\cite{vinyals2016matching,snell2017prototypical,sung2018learning,dong2018few,wang2019panet,wu2020meta,zhang2020sg,pambala2020sml} learned embeddings of training samples and test examples and a distance metric to measure the similarity between them.
\cite{vinyals2016matching} first introduced Matching Networks, which adopted a non-parametric principle by learning a differentiable K-Nearest Neighbour model. 
Prototypical Networks~\cite{snell2017prototypical} extended Matching Networks by producing a linear classifier instead of weighted nearest neighbor for each class.
Relation Networks~\cite{sung2018learning} further utilized a learnable non-linear comparator, instead of a fixed linear comparator to define the optimal distance metric. 
These methods relied on pre-trained embedding modules, which are trained on seen tasks, thus can not effectively handle unseen few-shot fashion landmark detection tasks, especially when there exists huge discrepancy between seen tasks and unseen tasks (\eg between detecting landmarks for upper-body clothes and for lower-body clothes). 

In the third category, the parameter prediction methods~\cite{bertinetto2016learning,shaban2017one,gidaris2019generating,gidaris2018dynamic,wang2019meta} learned to predict network parameters from a few annotated examples. 
It was first introduced in ~\cite{bertinetto2016learning}, which predicted the parameters of a network from a single representative of a class with a second neural network.
\cite{gidaris2018dynamic} enhanced the parameter prediction scheme with an attention based mechanism, which composes the classification weight vectors of unseen classes as a linear combination of those weight vectors of seen classes that are most similar to the few training examples.
Our method is most related to this category of meta-learning methods, which generates parameters for different number of landmark detectors of unseen tasks.
However, the previous work used a feed-forward network to obtain the parameters or simply fine-tuned the network with the predicted parameters, which limited the adaptation of the feature extraction network to unseen tasks.
In comparison, we further utilize meta-trained initialization parameters to learn a highly generalizable feature extraction network. 

\begin{figure*}[t]
	\vspace{-10pt}
	\begin{center}
		\includegraphics[width=0.9\linewidth]{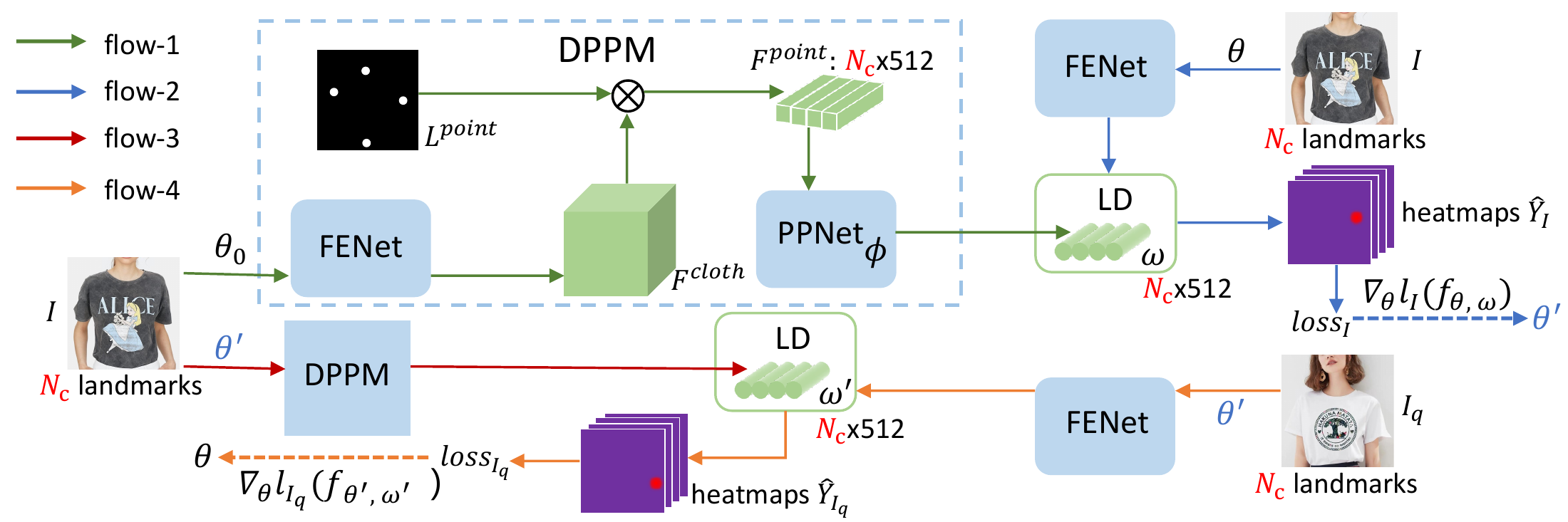}
	\end{center}
	\vspace{-10pt}
	\caption{The overall framework of \textbf{MetaCloth} during meta-training, which includes a feature extraction network (FENet), a parameter prediction network (PPNet) and landmark detectors (LD). FENet and LD constitute a fashion landmark detection model, where FENet extracts cloth-level features and LD predict landmark heatmaps. FENet and PPNet further make up a detector parameter prediction module (DPPM) to dynamically predict parameters for LD. For a task with $N_c$ landmarks, the \textbf{meta-training} contains four flows: \textbf{flow-1} adjusts the architecture of landmark detectors to predict $N_c$ landmarks through extracting $N_c$  landmark-level features and predicting parameters $\omega$ of $N_c$ landmark detectors on the support image $I$. \textbf{flow-2} tunes the feature extraction network to be task-specific, through updating the parameters $\theta$ of the FENet to ${\theta}^{\prime}$ with the loss calculated on $I$ using the predicted $\omega$. \textbf{flow-3} further updates the parameters of landmark detectors to be more suitable for this task through predicting parameters ${\omega}^{\prime}$ of $N_c$ landmark detectors with the updated ${\theta}^{\prime}$ on $I$. \textbf{flow-4} optimizes the parameters of the feature extraction network for initialization by updating the initial parameters $\theta$ of the FENet with the loss calculated on the query image $I_q$ using ${\theta}^{\prime}$ and ${\omega}^{\prime}$. During \textbf{meta-test}, the information flows are the same as those in the meta-training stage except that \textbf{flow-4} only includes a forward propagation to predict landmark heatmaps for a query image without the backward propagation in the dashed line.}
	\vspace{-10pt}
	\label{fig:meta}
\end{figure*}
\section{Method}
\subsection{Problem Setting} We introduce the few-shot dense fashion landmark detection in the regime of meta-learning, which trains a set of tasks, denoted as $\mathcal{T}=\{T_i^c\}$, where the $i^{th}$ task $T_i^c$ is adopted to train landmark detectors for the $c^{th}$ clothing category in a ``$N_c$-way $K$-shot'' setup. We have $c=1...C$ and a set of tasks are sampled for each $c$.
For $T_i^c$, we select $K$ images for the $c^{th}$ clothing category from the entire dataset, called a support set denoted as ${S_i^c}$. We have ${S_i^c}=\{(I_{ik}^c,Y_{ik}^c)\}_{k=1}^{K}$, where $I_{ik}^c$ is the $k^{th}$ image in the $i^{th}$ task of the $c^{th}$ category and $Y_{ik}^c$ represents a set of landmark locations in this image. The image $I_{ik}^c$ may contain $N_c$ landmarks.

The entire task set $\mathcal{T}$ is divided into two subsets without overlapping in terms of clothing categories, including $\mathcal{T}^{\mathrm{seen}}$ and $\mathcal{T}^{\mathrm{unseen}}$. $\mathcal{T}^{\mathrm{seen}}$ is used for meta-training, while $\mathcal{T}^{\mathrm{unseen}}$ is used for meta-test. In other words, given an image of an unseen category $c$ in meta-test, the meta-trained model should be able to predict $N_c$ landmarks with just $K$ images.

%
In meta-test, we perform episodes to evaluate the performance of the meta-trained model following~\cite{gidaris2019generating}, where each episode randomly samples a task from $\mathcal{T}^{\mathrm{unseen}}$.
Following the same setup in meta-training, each task is a ``$N_c$-way $K$-shot'' task  associated with a support set $S_i^c$ of $K$ images. The only difference is that $c$ is an unseen category.
Given $S_i^c$, the model is evaluated on a query set $Q_i^c$ to calculate error. The images of the support set and the query set do not have overlaps.
By following previous meta-learning work~\cite{finn2017model}, we report the errors by averaging the results of all the episodes.

\subsection{Overview}
In order to solve ``$N_c$-way $K$-shot'' fashion landmark detection, where $N_c$ varies across seen and unseen tasks, a model should possess two kinds of capabilities.
On one hand, the model needs to adaptively change the number of fashion landmarks that it can predict given a new task, where 
the parameter space can be adjusted dynamically.
On the other hand, the model needs to lean effective feature representations of fashion landmarks in a new task with a few annotated samples, so as to predict the locations of landmarks accurately.

To solve the above issues, this work proposes MetaCloth. It adaptively generates parameters for different number of landmark detectors when different tasks are presented, thus realizing a dynamic parameter space instead of fixing the architecture.
Once the parameters of the landmark detectors are predicted, it learns a generalizable feature extraction network from a few annotated samples with proper initialization, which is obtained through training a variety of tasks as prior knowledge. The learned feature extraction network extracts discriminative feature representations of fashion landmarks to facilitate the predictions of landmarks.

As shown in Fig.~\ref{fig:meta}, the meta-training stage of MetaCloth contains four flows represented by arrows of different colors. We design three components, including a feature extraction network (FENet), landmark detectors (LD) and a parameter prediction network (PPNet). In the following part, we introduce the information flows and the components in detail.

\subsection{Information Flows}
As shown in Fig.~\ref{fig:meta}, the meta-training stage of MetaCloth contains four flows. These flows successively predict the parameters of landmark detectors, tune the feature extraction network to be task-specific, update the parameters of landmark detectors, and optimize the  parameters of the feature extraction network, to obtain a set of good initialization parameters for efficient adaptation to unseen tasks.
\subsubsection{Flow-1} This flow adjusts the architecture of landmark detectors to adaptively change the number of landmarks that MetaCloth can predict, through predicting parameters for landmark detectors. 
Given a task with $N_c$ landmarks, this flow takes $N_c$ landmark-level features as inputs and generates parameters for $N_c$ landmark detectors.
Specifically, an image $I$ from $S_i^c$ with $N_c$ landmarks is fed into the base FENet to extract general cloth-level features $F^{\mathrm{cloth}}\in\mathbb{R}^{h\times w\times 512}$. 
The image $I$ has $N_c$ landmarks in the form of labelmaps $L^{\mathrm{point}}\in\mathbb{R}^{N_c\times h \times w}$ and each labelmap is a one-hot $h\times w$ binary mask, which indicates the location of a landmark.
The general landmark-level features $F^{\mathrm{point}}\in\mathbb{R}^{N_c\times 512}$ are extracted through a matrix multiplication between the labelmaps $L^{\mathrm{point}}$ and the cloth-level features $F^{\mathrm{cloth}}$. 
When there are multiple support images, we further average landmark-level features of all support images to obtain the final landmark-level features $F^{\mathrm{point}} \in\mathbb{R}^{N_c\times 512}$. $F^{\mathrm{point}}$ is then fed into PPNet to predict the parameters ${\omega}\in\mathbb{R}^{N_c\times 512}$ of LD. 
${\omega}$ has $N_c$ vectors with length 512, and each vector is used as a $1\times 1$ convolutional filter to detect a specific fashion landmark.
\subsubsection{Flow-2} %
This flow tunes the meta feature extraction network to be task-specific for the task ${T}_i^c$ from a few annotated samples.
Specifically, the image $I$ passes through the meta FENet with parameters ${\theta}$ and LD with the predicted parameters ${\omega}$ to estimate the landmark heatmaps ${\hat{Y}_I}\in\mathbb{R}^{N_c\times h \times w}$ on $I$ and calculate the loss ${\ell}_I$ between the predicted heatmaps and the ground-truth labelmaps. Then parameters ${\theta}$ of the meta FENet is optimized to ${\theta}^{\prime}$ using gradient descent as below:
\begin{equation}
	{\theta}^{\prime}\leftarrow{\theta} - {\beta}_1{\nabla}_{{\theta}}\ell_{I}(f_{{\theta},{\omega}})	
\end{equation}
where $f_{{\theta},{\omega}}$ consists of the meta FENet with parameters ${\theta}$ and LD with parameters ${\omega}$.

\subsubsection{Flow-3} %
This flow updates the parameters of the landmark detectors to be task-specific as well, with the updated meta feature extraction network, thus the landmark detectors are more suitable for the task ${T}_i^c$ to predict the landmarks.
Specifically, the image $I$ is fed into the meta FENet and PPNet again, which performs the same operations as those in the flow-1, with the updated parameters ${\theta}^{\prime}$ of the meta FENet after flow-2, to predict the parameters ${\omega}^{\prime}$ of LD. 
The parameters ${\omega}$ of LD in flow-1 are predicted from general landmark-level features extracted by the base FENet, to enable the tuning of the meta FENet. After the meta FENet is optimized from the given samples and extracts task-specific landmark-level features, we further update parameters ${\omega}^{\prime}$ of LD in flow-3 to make LD better adapt to the task ${T}_i^c$.

\subsubsection{Flow-4}  %
This flow optimizes for a set of good initialization parameters of the meta feature extraction network, such that it can be effectively adapted to unseen tasks with a few annotated samples when the parameters of the landmark detectors are predicted. 
During meta-training, a different image $I_q$ passes through the meta FENet with parameters ${\theta}^{\prime}$ and LD with the predicted parameters ${\omega}^{\prime}$ to estimate the landmark heatmap ${\hat{Y}_{I_q}}\in\mathbb{R}^{N_c\times h \times w}$ on $I_q$ and calculate the loss ${\ell}_{l_q}$. 
Instead of calculating the gradients of ${\theta}^{\prime}$ to optimize ${\theta}^{\prime}$, it calculates the gradients of the initial ${\theta}$ to optimize ${\theta}$ using gradient descents as below:
\begin{equation}
	{\theta}\leftarrow{\theta}-{\beta}_2{\nabla}_{{\theta}}\ell_{I_q}(f_{{\theta}^{\prime},{\omega}^{\prime}})
\end{equation}
where $f_{{\theta}^{\prime},{\omega}^{\prime}}$ consists of the meta FENet with parameters ${\theta}^{\prime}$ and LD with parameters ${\omega}^{\prime}$. Note that the optimization is performed over the model parameters ${\theta}$, whereas the loss function is computed using the updated model parameters ${\theta}^{\prime}$. 
The parameters of the meta feature extraction network are explicitly trained such that gradient descents with a small amount of training data from a new task will produce good generalization performance on that task when adopting the above parameter-prediction scheme.

During meta-test, this flow does not include the backward propagation to optimize parameters $\theta$. Instead, the query image $I_q$ passes through the meta FENet with parameters ${\theta}^{\prime}$ and LD with the predicted parameters ${\omega}^{\prime}$ to estimate the landmark heatmap ${\hat{Y}_{I_q}}$ as the final predictions.
\begin{algorithm}[h] 
	\caption{Meta-training of MetaCloth} 
	\label{alg:train} 
	\begin{algorithmic}[1] 
		\Require 
		task set $\mathcal{T}^{\mathrm{seen}}$
		\Require
		$\phi$: parameters of PPNet; ${\theta}_0$: parameters of the base FENet; learning rate ${\beta}_1$ and ${\beta}_2$
		\Ensure
		${\theta}$: parameters of the meta FENet\\
		Initialize ${\theta}$: parameters of the meta FENet
		\For{each task ${T}_i$ sampled from $\mathcal{T}^{\mathrm{seen}}$, ${T}_i$ consists of $S_i$ and $Q_i$}
		\State $F^{\mathrm{cloth}}$ = FENet($S_i$; ${\theta}_0$)
		\State $F^{\mathrm{point}}$ = $F^{\mathrm{cloth}} \times L^{\mathrm{point}}$
		\State ${\omega}$ = PPNet($F^{\mathrm{point}};\phi$)
		\State ${\theta}^{\prime}\leftarrow{\theta} - {\beta}_1{\nabla}_{{\theta}}\ell_{S_i}(f_{{\theta},{\omega}})$
		\State $F^{\mathrm{cloth}^{\prime}}$ = FENet($S_i$; ${\theta}^{\prime}$)
		\State $F^{\mathrm{point}^{\prime}}$ = $F^{\mathrm{cloth}^{\prime}} \times L^{\mathrm{point}}$
		\State ${\omega}^{\prime}$ = PPNet($F^{{\mathrm{point}}^{\prime}};\phi$)
		\State ${\theta}\leftarrow{\theta}-{\beta}_2{\nabla}_{{\theta}}\ell_{Q_i}(f_{{\theta}^{\prime},{\omega}^{\prime}})$
		\EndFor
	\end{algorithmic} 
\end{algorithm}

\begin{algorithm}[h] 
	\caption{Meta-test of MetaCloth} 
	\label{alg:test} 
	\begin{algorithmic}[1] 
		\Require 
		task set $\mathcal{T}^{\mathrm{unseen}}$
		\Require
		$\phi$: parameters of PPNet; ${\theta}_0$: parameters of the base FENet; ${\theta}$: parameters of the meta FENet; learning rate ${\beta}_1$
		\Ensure
		$\hat{Y}$: predicted landmarks
		\For{each task ${T}_i$ sampled from $\mathcal{T}^{\mathrm{unseen}}$, ${T}_i$ consists of $S_i$ and $Q_i$}
		\State $F^{\mathrm{cloth}}$ = FENet($S_i$; ${\theta}_0$)
		\State $F^{\mathrm{point}}$ = $F^{\mathrm{cloth}} \times L^{\mathrm{point}}$
		\State ${\omega}$ = PPNet($F^{\mathrm{point}};\phi$)
		\State ${\theta}^{\prime}\leftarrow{\theta} - {\beta}_1{\nabla}_{{\theta}}\ell_{S_i}(f_{{\theta},{\omega}})$
		\State $F^{\mathrm{cloth}^{\prime}}$ = FENet($S_i$; ${\theta}^{\prime}$)
		\State $F^{\mathrm{point}^{\prime}}$ = $F^{\mathrm{cloth}^{\prime}} \times L^{\mathrm{point}}$
		\State ${\omega}^{\prime}$ = PPNet($F^{{\mathrm{point}}^{\prime}};\phi$)
		\State $\hat{Y}$ = $f(Q_i;{{\theta}^{\prime},{\omega}^{\prime}})$
		\EndFor
	\end{algorithmic} 
\end{algorithm}

\subsubsection{Summary} Alg.~\ref{alg:train} and Alg.~\ref{alg:test} summary the complete information flows of meta-training and meta-test in MetaCloth. Note that multiple gradient updates are used to obtain ${\theta}^{\prime}$ in line 6 of Alg.~\ref{alg:train} and line 5 of Alg.~\ref{alg:test}. 
\subsection{Components}

\subsubsection{FENet}
The Feature Extraction Network (FENet) extracts cloth-level features of the images and serves two functions.
It first constitutes a complete fashion landmark detection model with landmark detectors, which can be tuned to a generalizable feature extraction network with proper initialization from a few annotated samples in a new task.
It further enables the implementation of a dynamic architecture, where the extracted landmark-level features are fed into the parameter prediction network to generate parameters for the corresponding number of landmark detectors.

As shown in Fig.~\ref{fig:meta}, FENet have two different set of parameters ${\theta}_0$ and $\theta$.
FENet with ${\theta}_0$ is a base feature extraction network, which is trained on all seen categories with standard supervised learning.
Specifically, suppose all seen categories define a total of $N_{all}$ landmarks. We train a supervised fashion landmark detection model $\mathcal{M}$, which consists of the base FENet with parameters ${\theta}_0$ and $N_{all}$ landmark detectors with parameters ${\omega}_0$, to detect $N_{all}$ landmarks.
After the training of the model $\mathcal{M}$, given an image $I$, the base FENet with parameters ${\theta}_0$ outputs a feature map $F^{\mathrm{cloth}}\in\mathbb{R}^{h\times w\times 512}$ and $N_{all}$ predicted labelmaps.
$N_c$ heatmaps belonging to the $c^{th}$ clothing category among the $N_{all}$ predicted heatmaps are selected as $L^{\mathrm{point}}$ to produce landmark-level features. During meta-test, ground-truth labelmaps are used.
Since the base FENet with parameters ${\theta}_0$ is trained to predict landmarks for all seen categories with conventional supervised learning, it extracts general landmark-level features.

FENet with ${\theta}$ is a meta feature extraction network, which is trained on a large number of tasks in the regime of meta-learning as shown in Alg.~\ref{alg:train}, for a set of initialization parameters that can be efficiently adapted to unseen tasks.

\subsubsection{LD} 
The Landmark Detectors (LD) perform pixel-wise classification on the cloth-level features and predict landmark heatmaps.
The architecture of LD can be adjusted adaptively to predict different number of fashion landmarks in different tasks, which is realized by a parameter prediction network to dynamically predict parameters for LD.
Specifically, for a task ${T}_i$ with $N_c$ landmarks, the parameters of LD have  $N_c$ vectors with length $512$, and each vector is used as a $1\times1$ convolutional filter to detect a specific fashion landmark. As shown in Fig.~\ref{fig:meta}, the parameters of LD are updated from $\omega$ to ${\omega}^{\prime}$ after the meta FENet is tuned, to make LD task-specific.

\begin{algorithm}[h] 
	\caption{The training of PPNet} 
	\label{alg:PPNet} 
	\begin{algorithmic}[1] 
		\Require 
		task set $\mathcal{T}^{\mathrm{seen}}$
		\Require
		${\theta}_0$: parameters of the base FENet; learning rate $\gamma$
		\Ensure
		$\phi$: parameters of PPNet\\
		Initialize $\phi$: parameters of PPNet
		\For{each task ${T}_i$ sampled from $\mathcal{T}^{\mathrm{seen}}$, ${T}_i$ consists of $S_i$ and $Q_i$}
		\State $F^{\mathrm{cloth}}$ = FENet($S_i$; ${\theta}_0$)
		\State $F^{\mathrm{point}}$ = $F^{\mathrm{cloth}} \times L^{\mathrm{point}}$
		\State ${\omega}$ = PPNet($F^{\mathrm{point}};\phi$)
		\State ${\phi}\leftarrow{\phi}-{\gamma}{\nabla}_{{\phi}}\ell_{Q_i}(f_{{\theta}_0,{\omega}})$
		\EndFor
	\end{algorithmic} 
\end{algorithm}

\subsubsection{PPNet}
The Parameter Prediction Network (PPNet) predicts parameters of the landmark detectors from landmark-level features. It enables the dynamic adjustment of the architecture to adaptively change the number of landmarks that MetaCloth can predict across tasks.
Specifically, taking $N_c$ landmark-level features as input, PPNet predicts $N_c$ vectors with length $512$ as the parameters of landmark detectors, and each vector is used as a $1\times1$ convolutional filter to detect a specific fashion landmark. In this way, MetaCloth can dynamically generate arbitrary number of parameters for unseen tasks with arbitrary number of landmarks, as long as a few samples with annotations are provided. 

PPNet with parameters $\phi$ is trained on a large number of tasks using meta-learning as shown in Alg.~\ref{alg:PPNet}, where $f_{{\theta}_0,{\omega}}$ consists of the base feature extraction network FENet with parameters ${\theta}_0$ and landmark detectors with parameters ${\omega}$. 
Specifically, we train a set of tasks  $\mathcal{T}=\{T_i\}$ and ${T}_i$ contains a support set $S_i$ and a query set $Q_i$.
For each ${T}_i$, the support set $S_i$ is fed into the base FENet with parameters ${\theta}_0$ to extract landmark-level features. PPNet predict parameters $\omega$ of the landmark detectors from the landmark-level features. Then the query set $Q_i$ passes through the base FENet with ${\theta}_0$ and landmark detectors with $\omega$ to predict landmark heatmaps.
The loss between predicted labelmaps and ground-truth labelmaps on $Q_i$ are calculated to optimize the parameters $\phi$ of PPNet.

\subsection{Analysis of Parameter Prediction Scheme}
In this part, we compare MetaCloth with the existing few-shot learning methods~\cite{hypernetworks, tail, analogy, motion, regression, gidaris2018dynamic, qiao2018few} that adopt parameter prediction scheme. These methods can be classified into two categories. The first category~\cite{regression, motion, tail} learns a network that transforms few-shot model parameters to many-shot parameters, while the second category~\cite{hypernetworks, qiao2018few, gidaris2018dynamic, analogy} learns a network that generates classification parameters from feature embeddings. Our method adopts the second category, which learns a parameter prediction network to predict parameters of landmark detectors from landmark-level features, since it can dynamically generate parameters for different numbers of landmarks. By contrast, learning a transformation 
between few-shot model parameters to many-shot parameters in the first category requires a fixed architecture, thus is less flexible in solving ``$N_c$-way $K$-shot'' tasks. Previous methods in the second category rely on a feature extraction network trained on seen tasks to obtain the features for parameter prediction, which limits the adaptation of the feature extraction network and landmark detectors to unseen tasks. In contrast, MetaCloth explicitly optimizes the feature extraction network for initialization through the parameter prediction scheme. Since the parameters of landmark detectors are correlated with the parameters of the feature extraction network, such optimization leads to a set of good initialization parameters that can be tuned to unseen tasks with stronger feature representations and more effective landmark positioning.

\section{Experiments}
\subsection{Dataset}
Among all fashion datasets, DeepFashion2~\cite{ge2019deepfashion2} contains the most abundant clothing landmark annotations, which covers comprehensive clothing categories.
Specifically, DeepFashion2 defines a total of 294 clothing landmarks on 13 clothing categories, where each clothing category is defined with different fashion landmarks and the number of defined landmarks ranges from 8 to 39 among 13 categories.
We use DeepFashion2 to evaluate the few-shot fashion landmark detection.

Following the construction of few-shot object detection benchmarks in \cite{kang2019few}, we build four few-shot fashion landmark detection benchmarks.
In each benchmark, out of the 13 clothing categories in DeepFashion2, we select 6 clothing categories with images in the training set to construct seen tasks as $\mathcal{T}^{\mathrm{seen}}$ and keep the remaining 7 categories with images in the validation set to construct unseen tasks as $\mathcal{T}^{\mathrm{unseen}}$, where each task in a ``$N_c$-way $K$-shot'' setup samples $K$ images of category $c$ with $N_c$ landmarks from DeepFashion2. Different benchmarks have different $\mathcal{T}^{\mathrm{seen}}$/$\mathcal{T}^{\mathrm{unseen}}$ splits.  

Benchmark-1 explores how the few-shot fashion landmark detection models perform when transferring from half-body clothing categories to full-body clothing categories by setting only upper-body and lower-body clothing categories in $\mathcal{T}^{\mathrm{seen}}$ and all the full-body clothing categories in  $\mathcal{T}^{\mathrm{unseen}}$.

Benchmark-2 investigates the few-shot fashion landmark detection models' ability to transfer from upper-body clothing categories to lower-body clothing categories by setting only upper-body clothing categories in $\mathcal{T}^{\mathrm{seen}}$. In this way, none of lower-body clothing categories in $\mathcal{T}^{\mathrm{unseen}}$ share similar structure with those clothing categories in $\mathcal{T}^{\mathrm{seen}}$.

Benchmark-3 evaluates the few-shot fashion landmark detection models' performance to transfer from clothing categories with fewer landmarks (\eg 8 landmarks for ``skirt'') to clothing categories with more landmarks (\eg 39 landmarks for ``long sleeve outwear''). Specifically, clothing categories are sorted by the number of landmarks in descending order and $\mathcal{T}^{\mathrm{seen}}$ selects the last six categories.

Benchmark-4 studies the performance of the few-shot fashion landmark detection models under a random split scenario while the other three benchmarks are constructed to mimic specific transfer scenarios. The 6 clothing categories in $\mathcal{T}^{\mathrm{seen}}$ are randomly selected from the 13 clothing categories.

During meta-test, we sample 100 random episodes for each clothing category from $\mathcal{T}^{\mathrm{unseen}}$ and a total of 700 episodes are constructed. 
Specifically, each episode samples a $N_c$-way K-shot tasks and 24 query images will be evaluated for the task.
We adopt normalized error (NE)~\cite{yu2019layout} as evaluation metric, which indicates better performance with smaller value.
NE is defined as the $l_2$ distance between predicted landmarks and ground truth landmarks in the normalized coordinate space(i.e. divided by the square root of the clothing area).
Results over 700 episodes are averaged to evaluate models.

\subsection{Implementation Details}
MetaCloth is made up of a feature extraction network (FENet), landmark detectors (LD) and a parameter prediction network (PPNet).
To focus on the transferability of MetaCloth to unseen tasks, we adopt simple architectures for each component.
FENet employs ResNet-50~\cite{he2016deep} pretrained from ImageNet~\cite{deng2009imagenet} as the network backbone, followed by ten convolution layers and two deconvolution layers in reference with~\cite{he2017real}.
LD is made up of $1\times 1$ convolutional filters and PPNet consists of two fully-connected layers.
For training MetaCloth, taking benchmark-1 as an example with a total of 112 defined landmarks in $\mathcal{T}^{\mathrm{seen}}$, we introduce the details.
We first resize each clothing item of DeepFashion2 images to $384\times 384$.
We train the supervised landmark detection model $\mathcal{M}$, which consists of the base FENet with parameters ${\theta}_0$ and LD with parameters ${\omega}_0$ on 112 defined landmarks for 20 epochs starting from a learning rate of 0.01 and reducing it by 10 at 10 and 15 epochs.
PPNet is meta-trained with 40,000 tasks, starting from a learning rate 0.002 and reducing it by 10 at 20000 tasks. 
The meta FENet is meta-trained for 40,000 tasks. Each task is trained using 12 gradient steps with learning rate ${\beta}_1=0.01$ and learning rate ${\beta}_2=0.0002$.

\subsection{Baselines}
Few-shot dense fashion landmark detection is a new problem and hasn't been tackled before. 	We tailor the existing few-shot learning methods to address the proposed task and serve as eight baseline.
These baselines employ the same network architecture as the landmark detection model of MetaCloth, which includes a feature extraction network and landmark detectors with $1\times 1$ convolution.

\subsubsection{Fine-tune (FT)} During training, parameters of the feature extraction network and landmark detectors are optimized on all seen categories with conventional supervised learning. During test, for a $N_c$-Way K-shot task, it tunes the pre-trained feature extraction network and trains parameters of $N_c$ landmark detectors, which adopts the same practice as the baseline model in previous few-shot learning work~\cite{qiao2018few}. 

\subsubsection{MAML~\cite{fink2004object}} The existing MAML requires fixed number of parameters in a model, thus cannot be directly applied to solve few-shot fashion landmark detection, where different tasks have different number of landmark detectors. We adjust it by setting the number of landmark detectors as the maximum number of landmarks in all tasks, $\ie$ 39 in DeepFashion2, and correspondingly expanding the number of landmarks per clothes image to 39, guaranteeing that each of the 39 detectors is semantically meaningful during meta-training.

\subsubsection{Few-shot Weight Generator (WG)~\cite{gidaris2018dynamic}} The original WG solves few-shot image classification by predicting classification weights for unseen categories from the feature vectors of the training examples.
We adjust WG to solve the few-shot fashion landmark detection by predicting the parameters of landmark detectors from the landmark-level feature vectors of the training examples.

\subsubsection{WG-ATT~\cite{gidaris2018dynamic}} Besides generating the parameters from the features as WG, this method further adopts an attention-based weight composition mechanism, where the classification weights
of a novel category are composed as a linear combination of those base classification weight vectors. We adjust WG-ATT through generating the parameters of landmark detectors both from the landmark-level features and the parameters of landmark detectors from other clothing categories.

\subsubsection{Prototypical Networks (PROTO)~\cite{snell2017prototypical}} Prototypical Networks construct a prototype for each class and classifies each query example as the class whose prototype is ``nearest'' to it. We construct a prototype for each landmark with the landmark-level features and assign the location of a query image that is ``nearest''to the prototype as the landmark location.

\subsubsection{PROTO-MA~\cite{metadataset}} This method combines Prototypical Networks and MAML. When updating the parameters, the gradients are allowed to flow through the Prototypical Network-equivalent linear layer initialization. 

\subsubsection{Model Regression Network (MRN)~\cite{regression}} This method trains a deep regression network to learn a generic, category agnostic transformation from models learned from a few samples to models learned from large samples, and uses this transformation in learning models for novel categories.

\subsubsection{Proactive and Adaptive Meta-learning (PAML)~\cite{motion}} This method produces a generic initial model through aggregating con-textual information from a variety of tasks, while effectively learns how to transform few-shot model parameters to many-shot model parameters. 

\begin{table*}\small\centering
	\vspace{-10pt}
	\caption{Few-shot fashion landmark detection results on four benchmarks. The evaluation metric is normalized error (NE) with 95\% confidence intervals ($\times{10}^{-2}$), which is averaged over 700 episodes and \textbf{smaller value indicates better performance}. Results on different shots are further averaged as the mean value. The best performance of each benchmark is bold.} 
	\scalebox{0.9}{
		\begin{tabular}{cc|cccccccc|c}
			\toprule[1.5pt]
			&Shot&FT&MAML&WG&WG-ATT&PROTO&PROTO-MA&MRN&PAML&Ours\\
			\toprule[1.5pt]
			\multirow{5}*{B-1}&1&0.266$\pm$0.8&0.212$\pm$1.2&0.104$\pm$2.2&0.103$\pm$2.2&0.109$\pm$1.3&0.126$\pm$1.6&0.320$\pm$1.0&0.165$\pm$2.7&\textbf{0.086}$\pm$1.4\\
			&3&0.177$\pm$0.6&0.130$\pm$1.0&0.088$\pm$2.0&0.086$\pm$1.9&0.085$\pm$1.1&0.081$\pm$1.2&0.229$\pm$0.7&0.101$\pm$1.8&\textbf{0.081}$\pm$1.1\\
			&5&0.154$\pm$0.5&0.106$\pm$0.9&0.086$\pm$1.9&0.084$\pm$1.9&0.070$\pm$0.9&0.067$\pm$1.2&0.209$\pm$0.5&0.088$\pm$1.5&\textbf{0.065}$\pm$0.9\\
			&8&0.140$\pm$0.4&0.090$\pm$0.9&0.085$\pm$1.9&0.082$\pm$1.8&0.068$\pm$0.8&0.062$\pm$1.0&0.189$\pm$0.5&0.078$\pm$1.3&\textbf{0.057}$\pm$0.8\\
			&10&0.133$\pm$0.4&0.088$\pm$0.7&0.084$\pm$1.8&0.081$\pm$1.6&0.067$\pm$0.8&0.059$\pm$0.9&0.185$\pm$0.3&0.076$\pm$1.1&\textbf{0.055}$\pm$0.6\\
			&Mean&0.174&0.125&0.089&0.087&0.080&0.079&0.226&0.102&\textbf{0.069}\\
			\bottomrule[1.5pt]
			\multirow{5}*{B-2}
			&1&0.257$\pm$0.9&0.215$\pm$2.3&0.186$\pm$5.2&0.185$\pm$5.1&0.179$\pm$4.2&0.187$\pm$4.3&0.360$\pm$2.4&0.233$\pm$4.9&\textbf{0.145}$\pm$2.1\\
			&3&0.186$\pm$0.7&0.157$\pm$2.0&0.175$\pm$5.0&0.164$\pm$4.8&0.145$\pm$4.2&0.123$\pm$2.5&0.293$\pm$2.3&0.153$\pm$3.7&\textbf{0.121}$\pm$1.7\\
			&5&0.165$\pm$0.6&0.126$\pm$1.7&0.173$\pm$4.9&0.160$\pm$4.7&0.135$\pm$3.9&0.103$\pm$2.3&0.252$\pm$1.5&0.139$\pm$3.5&\textbf{0.096}$\pm$1.5\\
			&8&0.148$\pm$0.6&0.106$\pm$1.6&0.172$\pm$4.9&0.158$\pm$4.5&0.128$\pm$3.7&0.094$\pm$2.1&0.239$\pm$1.4&0.131$\pm$3.4&\textbf{0.080}$\pm$1.4\\
			&10&0.142$\pm$0.5&0.100$\pm$1.5&0.170$\pm$4.7&0.157$\pm$4.5&0.127$\pm$3.6&0.082$\pm$1.7&0.236$\pm$1.2&0.129$\pm$3.1&\textbf{0.077}$\pm$1.2\\
			&Mean&0.180 &0.141&0.175&0.165&0.143&0.118&0.276&0.157 &\textbf{0.104}\\
			\bottomrule[1.5pt]
			\multirow{5}*{B-3}
			&1&0.312$\pm$1.5&0.241$\pm$2.0&0.163$\pm$2.3&0.150$\pm$2.1&0.150$\pm$1.9&0.150$\pm$1.9&0.325$\pm$2.0&0.222$\pm$1.6&\textbf{0.141}$\pm$1.9\\
			&3&0.241$\pm$1.2&0.175$\pm$1.7&0.145$\pm$2.2&0.127$\pm$1.9&0.114$\pm$1.6&0.103$\pm$1.7&0.230$\pm$1.9&0.131$\pm$1.4&\textbf{0.099}$\pm$1.5\\
			&5&0.217$\pm$1.1&0.155$\pm$1.5&0.144$\pm$2.0&0.124$\pm$1.8&0.106$\pm$1.5&0.093$\pm$1.6&0.204$\pm$1.8&0.115$\pm$1.4&\textbf{0.088}$\pm$1.3\\
			&8&0.201$\pm$0.8&0.147$\pm$1.4&0.143$\pm$1.9&0.122$\pm$1.7&0.102$\pm$1.4&0.090$\pm$1.3&0.185$\pm$1.8&0.106$\pm$1.3&\textbf{0.083}$\pm$1.2\\
			&10&0.189$\pm$0.6&0.146$\pm$1.1&0.142$\pm$1.9&0.121$\pm$1.7&0.097$\pm$1.2&0.081$\pm$1.2&0.181$\pm$1.7&0.103$\pm$1.2&\textbf{0.081}$\pm$1.0\\
			&Mean&0.232&0.173&0.147&0.129&0.114&0.103&0.225&0.135&\textbf{0.098}\\
			\bottomrule[1.5pt]
			\multirow{5}*{B-4}
			&1&0.222$\pm$1.7&0.183$\pm$3.0&0.144$\pm$5.3&0.142$\pm$5.2&0.145$\pm$4.1&0.143$\pm$4.0&0.333$\pm$2.1&0.192$\pm$5.2&\textbf{0.138}$\pm$2.3\\
			&3&0.168$\pm$1.6&0.141$\pm$2.0&0.121$\pm$4.7&0.113$\pm$4.4&0.115$\pm$3.6&\textbf{0.105}$\pm$3.0&0.250$\pm$1.4&0.117$\pm$3.5&0.113$\pm$1.9\\
			&5&0.153$\pm$1.5&0.110$\pm$1.7&0.113$\pm$4.5&0.108$\pm$4.2&0.103$\pm$3.3&0.090$\pm$2.5&0.222$\pm$1.3&0.098$\pm$2.9&\textbf{0.089}$\pm$1.6\\
			&8&0.145$\pm$1.4&0.095$\pm$1.5&0.113$\pm$4.4&0.108$\pm$4.2&0.098$\pm$3.2&0.078$\pm$2.1&0.204$\pm$0.9&0.087$\pm$2.5&\textbf{0.077}$\pm$1.4\\
			&10&0.141$\pm$1.2&0.089$\pm$1.4&0.112$\pm$4.2&0.107$\pm$4.1&0.096$\pm$2.9&0.073$\pm$1.9&0.200$\pm$0.8&0.084$\pm$2.4&\textbf{0.069}$\pm$1.3\\
			&Mean&0.166&0.124&0.121&0.116&0.111&0.098&0.242&0.116&\textbf{0.097}\\
			\bottomrule[1.5pt]			
	\end{tabular}}
	\vspace{-10pt}
	\label{tab:performance}
\end{table*}

\subsection{Comparisons with Baselines} \label{sec:few-shot}
\subsubsection{Overall Performance}
We present the few-shot landmark detection results on four benchmarks in Table~\ref{tab:performance}.
First of all, MetaCloth outperforms all the baselines in all benchmarks on the averaged results. MetaCloth achieves the smallest errors on three benchmarks with specific transfer scenarios on all different shots, including transferring from half-body clothing categories to full-body clothing categories in benchmark-1, from upper-body clothing categories to lower-body clothing categories in benchmark-2, and from clothing categories with fewer landmarks to clothing categories with more landmarks in benchmark-3. In benchmark-4 under a random split scenario, MetaCloth achieves the best performance except for shot-3. We can conclude that MetaCloth is a highly effective framework for handling $N_c$-way $K$-shot fashion landmark detection tasks.

Furthermore, WG and WG-ATT both adopt the parameter-prediction scheme, where the parameters of the landmark detectors are generated from the landmark-level features. However, their performances lag far behind our MetaCloth in all benchmarks. When there exists huge discrepancy between the seen clothing categories and the unseen categories (\ie from upper-body clothing categories to lower-body clothing categories in benchmark-2), they fail to generate meaningful parameters from landmark-level features with a fixed feature extractor trained on seen categories. By contrast, besides dynamically generating parameters, MetaCloth learns a highly generalizable feature extraction network with meta-learned initialization, such that it can be better adapted to unseen tasks with a few annotated samples.

\begin{figure}[t]
	\begin{center}
		\includegraphics[width=1\linewidth]{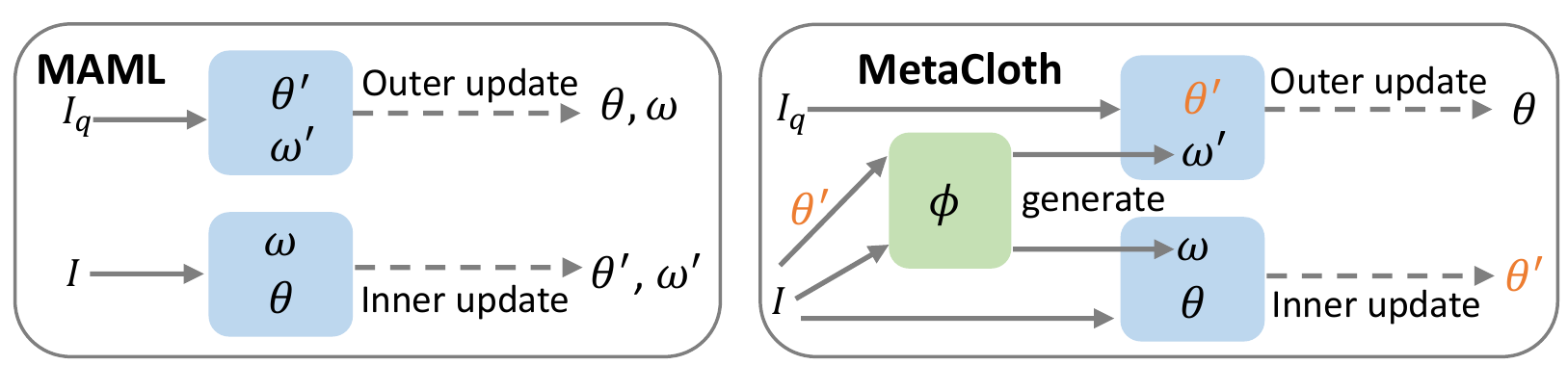}
	\end{center}
	\vspace{-10pt}
	\caption{The high-level diagram of MAML~\cite{finn2017model} and MetaCloth. $\theta$, $\omega$ and $\phi$ represents the parameters of the feature extraction network, landmark detectors and the parameter prediction network respectively.}
	\vspace{-10pt}
	\label{fig:comparison}
\end{figure}

Finally, when comparing between FT and MAML, PROTO and PROTO-MA, MRN and PAML, the latter baselines perform better than the former baselines in all benchmarks. Since the latter baselines including MAML, PROTO-MA and PAML all train a variety of tasks to optimize the initialization of the models, these models can better tune their parameters for unseen clothing categories. However, these MAML-based methods all drop behind our MetaCloth. MAML jointly meta-optimizes the feature extraction network and landmark detectors as shown in Fig.~\ref{fig:comparison}. As pointed by ~\cite{motion}, plain updates can only slightly modify its parameters, otherwise, it would lead to severe over-fitting to the new data. PROTO-MA initializes the task-specific linear landmark detectors from the Prototypical Network-equivalent weights and bias. However, it uses a pre-defined fixed metric (\ie Euclidean distance), which assumes linear separability after the landmark features, and hence is totally limited by the efficacy of the feature extraction network. PAML learns the initialization of a model and meanwhile learns a transformation from few-shot model parameters to many-shot model parameters. The assumption that there exists a generic non-linear transformation in the model parameter space is not proved in a dynamic parameter space to solve the few-shot landmark detection problem.
In contrast, while MetaCloth also optimizes for a set of good initialization parameters through learning abundant tasks, it adaptively generate parameters for different number of landmark detectors from the landmark-level features and accordingly optimizes the feature extraction network as summarized in Fig.~\ref{fig:comparison}. Since the parameters of landmark detectors are correlated with the parameters of the feature extraction network through a parameter prediction network, such optimization leads to not only stronger feature representations but also more effective landmark positioning, and at the same time realizes a dynamic parameter space. The exceeding performance of MetaCloth over these baseline models shows the superiority of our method in tuning the parameters with the parameter-prediction scheme for better adaptation to unseen tasks.

\begin{table*}\small\centering
	\caption{Few-shot fashion landmark detection results on three benchmarks of different unseen categories with 8 shots. The evaluation metric is normalized error (NE) with 95\% confidence intervals ($\times{10}^{-2}$) and \textbf{smaller value indicates better performance}. The best performance of each category is bold.}
	\scalebox{0.90}{
		\begin{tabular}{c|cccccccc|c}
			\toprule[1.5pt]
			Benchmark-1&FT&MAML&WG&WG-ATT&PROTO&PROTO-MA&MRN&PAML&Ours\\
			\toprule[1.5pt]
			long sleeve top&0.156$\pm$0.6&0.101$\pm$2.0&0.045$\pm$0.6&\textbf{0.043}$\pm$1.2&0.055$\pm$1.7&0.050$\pm$2.1&0.189$\pm$0.9&0.063$\pm$1.9&0.053$\pm$1.5\\
			short sleeve outwear&0.140$\pm$0.8&0.082$\pm$2.1&0.049$\pm$0.8&\textbf{0.043}$\pm$1.6&0.048$\pm$1.9&0.045$\pm$1.4&0.173$\pm$1.1&0.059$\pm$1.3&0.047$\pm$1.0\\
			trousers&0.162$\pm$1.2&0.112$\pm$2.5&0.190$\pm$4.2&0.182$\pm$4.9&0.101$\pm$2.2&0.094$\pm$2.9&0.193$\pm$1.5&0.135$\pm$3.6&\textbf{0.086}$\pm$2.1\\
			short sleeve dress&0.109$\pm$1.1&0.067$\pm$1.3&0.051$\pm$1.0&0.052$\pm$1.6&0.048$\pm$1.0&0.045$\pm$1.7&0.182$\pm$1.6&0.051$\pm$1.5&\textbf{0.042}$\pm$1.3\\
			long sleeve dress&0.152$\pm$1.3&0.102$\pm$2.5&0.054$\pm$1.1&0.054$\pm$1.8&0.059$\pm$2.2&0.054$\pm$2.0&0.190$\pm$1.8&0.062$\pm$1.7&\textbf{0.051}$\pm$1.2\\
			vest dress&0.106$\pm$1.2&0.059$\pm$1.4&0.076$\pm$2.4&0.076$\pm$2.0&0.057$\pm$2.2&0.054$\pm$1.7&0.191$\pm$1.7&0.068$\pm$1.7&\textbf{0.046}$\pm$1.3\\
			sling dress&0.142$\pm$1.4&0.098$\pm$2.6&0.129$\pm$3.5&0.127$\pm$4.3&0.108$\pm$3.4&0.091$\pm$3.1&0.206$\pm$1.9&0.106$\pm$2.9&\textbf{0.075}$\pm$1.9\\
			\toprule[1.5pt]
			Benchmark-2&FT&MAML&WG&WG-ATT&PROTO&PROTO-MA&MRN&PAML&Ours\\
			\toprule[1.5pt]
			shorts&0.192$\pm$2.4&0.144$\pm$4.0&0.313$\pm$4.2&0.272$\pm$3.0&0.195$\pm$2.1&0.149$\pm$2.9&0.271$\pm$3.1&0.206$\pm$3.8&\textbf{0.119}$\pm$2.9\\
			trousers&0.158$\pm$2.3&0.122$\pm$3.8&0.286$\pm$3.1&0.266$\pm$2.3&0.225$\pm$3.5&0.134$\pm$2.3&0.276$\pm$4.4&0.204$\pm$3.4&\textbf{0.097}$\pm$2.3\\
			skirt&0.196$\pm$2.5&0.152$\pm$4.2&0.299$\pm$3.8&0.272$\pm$2.5&0.206$\pm$3.9&0.145$\pm$2.2&0.276$\pm$4.3&0.228$\pm$4.5&\textbf{0.121}$\pm$3.5\\
			short sleeve dress&0.108$\pm$1.2&0.062$\pm$1.8&0.051$\pm$1.9&0.051$\pm$1.1&0.048$\pm$1.7&0.043$\pm$1.2&0.195$\pm$3.5&0.049$\pm$1.2&\textbf{0.042}$\pm$1.1\\
			long sleeve dress&0.136$\pm$1.6&0.104$\pm$3.1&0.052$\pm$2.7&0.048$\pm$1.9&0.049$\pm$1.4&\textbf{0.043}$\pm$1.5&0.201$\pm$3.9&0.052$\pm$1.8&0.050$\pm$1.4\\
			vest dress&0.104$\pm$1.3&0.055$\pm$1.0&0.077$\pm$2.0&0.076$\pm$1.8&0.062$\pm$1.0&0.054$\pm$1.3&0.228$\pm$3.7&0.071$\pm$1.7&\textbf{0.050}$\pm$1.6\\
			sling dress&0.140$\pm$1.8&0.103$\pm$3.5&0.126$\pm$2.9&0.120$\pm$2.1&0.110$\pm$2.3&0.089$\pm$2.6&0.228$\pm$4.0&0.107$\pm$3.1&\textbf{0.083}$\pm$2.1\\
			\toprule[1.5pt]
			Benchmark-3&FT&MAML&WG&WG-ATT&PROTO&PROTO-MA&MRN&PAML&Ours\\
			\toprule[1.5pt]
			short sleeve top&0.130$\pm$1.1&0.082$\pm$1.7&0.142$\pm$3.0&0.108$\pm$2.5&0.066$\pm$2.0&0.060$\pm$1.2&0.163$\pm$3.4&0.079$\pm$2.0&\textbf{0.056}$\pm$1.2\\
			long sleeve top&0.234$\pm$2.6&0.188$\pm$3.1&0.206$\pm$4.5&0.176$\pm$3.7&0.135$\pm$3.3&0.120$\pm$3.0&0.233$\pm$4.1&0.137$\pm$2.8&\textbf{0.118}$\pm$2.7\\
			short sleeve outwear&0.208$\pm$2.9&0.170$\pm$3.7&0.167$\pm$3.9&0.140$\pm$3.5&0.112$\pm$3.9&0.100$\pm$2.5&0.199$\pm$3.6&0.122$\pm$3.6&\textbf{0.092}$\pm$2.3\\
			long sleeve outwear&0.263$\pm$3.3&0.238$\pm$3.9&0.184$\pm$4.4&0.161$\pm$4.0&0.144$\pm$4.2&0.133$\pm$4.0&0.238$\pm$4.5&0.157$\pm$4.2&\textbf{0.117}$\pm$3.8\\
			short sleeve dress&0.166$\pm$2.2&0.095$\pm$2.7&0.083$\pm$2.1&0.072$\pm$1.7&0.062$\pm$2.0&0.060$\pm$1.8&0.138$\pm$3.2&0.072$\pm$1.7&\textbf{0.053}$\pm$1.7\\
			long sleeve dress&0.241$\pm$3.4&0.170$\pm$3.1&0.137$\pm$3.3&0.117$\pm$2.9&0.101$\pm$3.5&0.088$\pm$2.0&0.190$\pm$4.3&0.115$\pm$2.1&\textbf{0.086}$\pm$1.9\\
			sling dress&0.164$\pm$2.6&0.086$\pm$2.3&0.079$\pm$2.8&0.077$\pm$2.2&0.092$\pm$2.7&0.066$\pm$1.3&0.131$\pm$3.5&0.059$\pm$1.4&\textbf{0.056}$\pm$1.2\\
			\bottomrule[1.5pt]
	\end{tabular}}
	\vspace{-10pt}
	\label{tab:seperate}
\end{table*}

\subsubsection{Comparison between Different Benchmarks} We further have several observations from Table~\ref{tab:performance} when comparing the performance of the models between the first three benchmarks, which are constructed to mimic the specific transfer scenarios.

First of all, generally speaking, in benchmark-1, all models have smaller errors compared with the other two benchmarks, which are trained on half-body clothing categories and evaluated on full-body clothing categories. Since the full-body clothes can be regarded as the combination of a upper-body clothes and a low-body clothes, the models have less difficulty in detecting fashion landmarks for the full-body clothes. In this benchmark, MetaCloth achieves better performance than the baseline models.

Second, in benchmark-2, all the baselines except FT and MAML show the worst performance among the three benchmarks, which transfers from upper-body clothing categories to lower-body clothing categories.  In this benchmark, the clothing categories in unseen tasks have totally different appearance and structure from the clothing categories in seen tasks. WG, WG-ATT and PROTO use the trained feature extraction network on seen tasks and can not extract effective feature representations for unseen tasks. PROTO-MA and PAML tune the feature extraction network from the meta-learned model initialization with a few samples. However, PROTO-MA is solely relied on the tuned feature extraction network with a pre-defined fixed metric while PAML subjects to the transformation from few-shot model parameters to many-shot model parameters learned from seen categories. 

\begin{figure*}[t]
	\vspace{10pt}
	\begin{center}
		\includegraphics[width=1.0\linewidth]{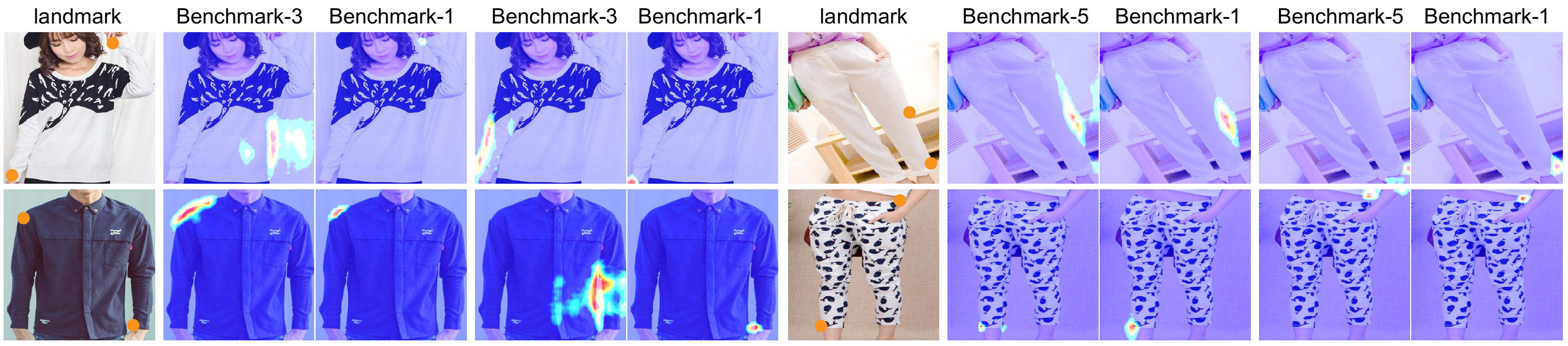}
	\end{center}
	\vspace{-10pt}
	\caption{The visualization of the predicted landmark heatmaps on MetaCloth in different benchmarks, where the models are trained with 8 annotated samples.}
	\vspace{-10pt}
	\label{fig:benchmarks}
\end{figure*}

Furthermore, in benchmark-3, FT and MAML have the largest error among all benchmarks, which learns unseen tasks with more landmarks from seen tasks with fewer landmarks.  In this benchmark, the parameter space of the models has huge discrepancy between the seen tasks and the unseen tasks, due to the difference between the number of landmark detectors. FT and MAML both jointly fine-tune the feature extraction network and landmark detectors when learning unseen tasks, which can overfit to a few annotated samples. The poor performances demonstrate that they can not effectively adjust their parameter space in a new task. 

Finally, by comparison, in benchmark-2 and benchmark-3, MetaCloth still achieves better results compared with the baseline models, proving its superiority in adjusting the parameter space and learning a effective feature extractor even in difficult scenarios. To visually show the influence of different benchmarks on the transferability of our model, we visualize the heatmaps of the same clothing category in different benchmarks predicted by MetaCloth in Fig.~\ref{fig:benchmarks}, where the ``landmark'' column presents the ground-truth landmark locations. We can see that for the clothing category ``long sleeve shirt‘’ and ``trousers'',  the landmark heatmaps predicted by MetaCloth in benchmark-1 are bright in the region of landmarks, which accurately indicate the locations of the landmarks without ambiguity. By contrast, MetaCloth in benchmark-3 predicts messy heatmaps for landmarks on the edge of the sleeves and MetaCloth in benchmark-5 causes confusion in the predicted locations of landmarks on the trouser legs. In benchmark-3, all the seen clothing categories are sleeveless, thus it is more difficult for the model to learn to detect the landmarks on the sleeves with only a few samples. In benchmark-5, the model is trained only on upper-body clothes, which bring a severe challenge to it to transfer to the lower-body clothes.

\subsubsection{Transferability to Different Clothing Categories} 
\begin{figure}[t]
	\vspace{-10pt}
	\begin{center}
		\includegraphics[width=0.8\linewidth]{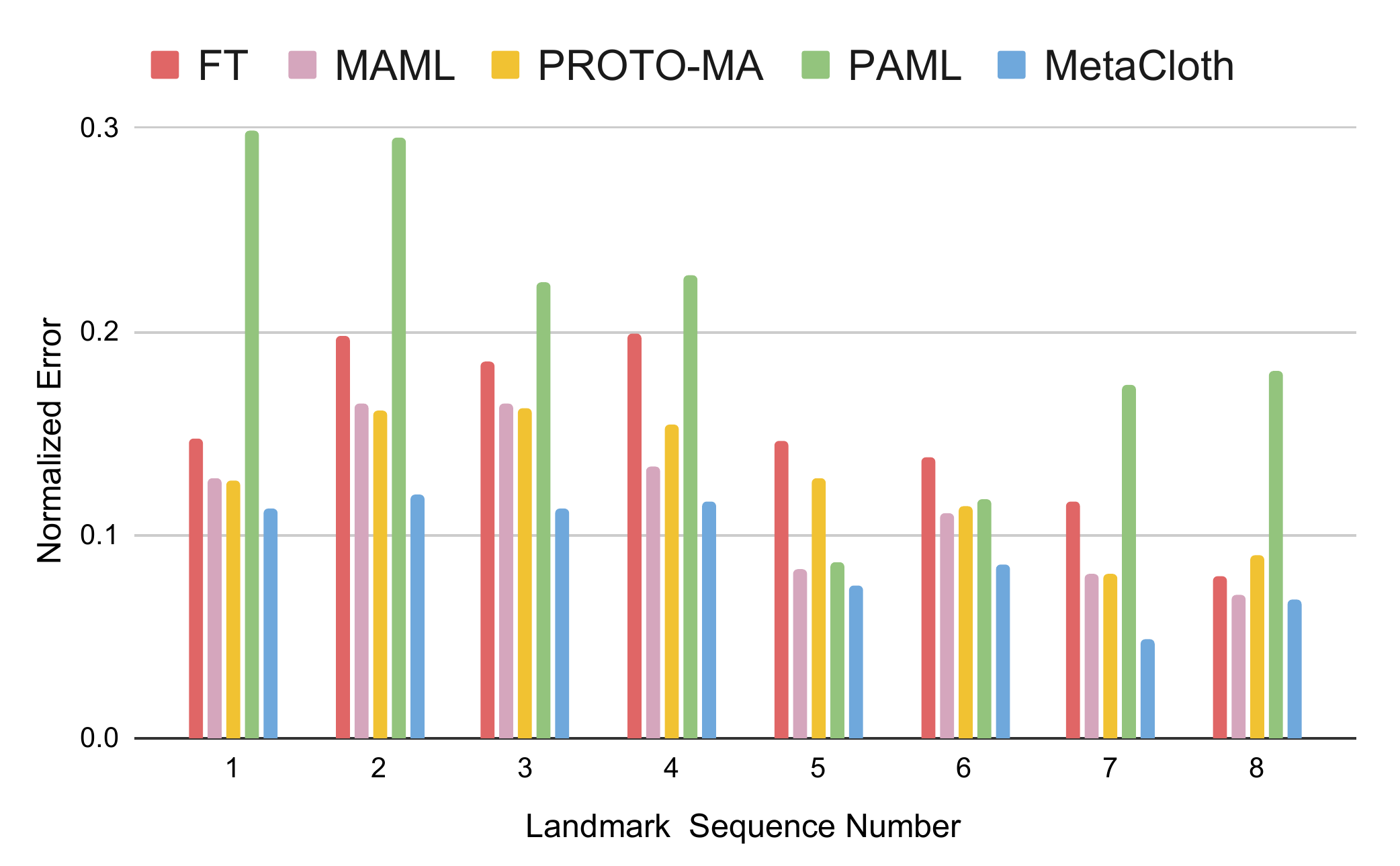}
	\end{center}
	\vspace{-10pt}
	\caption{Few-shot fashion landmark detection results of individual fashion landmarks of the clothing category ``trousers'' in benchmark-2 with 8 shots. The evaluation metric is normalized error (NE) and smaller value is better.}
	\vspace{-10pt}
	\label{fig:landmark}
\end{figure}

Table~\ref{tab:seperate} further presents separate results on different unseen categories in three benchmarks with specific transfer scenarios. The number of annotated samples for each unseen task is 8.
Overall, MetaCloth outperforms baselines on all unseen categories in different benchmarks, which indicates its advantage in transferring to the arbitrary clothing category in different scenarios.  
Furthermore, the performances of the models vary when they detect landmarks for different clothing categories.
For example, in benchmark-1, the baseline models have larger error on ``trousers''. The mean error on ``trousers'' in WG and WG-ATT is even several times larger than the other clothing categories. There is only one lower-body clothing category in training, which leads to the difficulty of learning effective feature representations for the lower-body clothing category ``trousers'' with only a few annotated samples.
In benchmark-2, the baseline models have significantly larger error on the lower-body clothing categories such as ``shorts'', ``trousers'' and ``skirt''. The lower-body clothing categories in unseen tasks have totally different appearance and structures with those in seen tasks, thus bring an extremely serve challenge to solve the few-shot fashion landmark detection. We calculate the landmark detection results of individual landmarks of the clothing category ``trousers'' in Fig.~\ref{fig:landmark}. For the symmetry landmarks, we average the normalized error and show the results of 9 landmarks. As can be seen from Fig.~\ref{fig:landmark}, MetaCloth achieves the smallest error on each individual landmark of ``trousers''.
In benchmark-3,  baseline model have larger error on clothing categories with more landmarks (\eg ``long sleeve top'' with 33 landmarks and ``long sleeve outwear'' with 39 landmarks). Since the clothing categories in training have fewer landmarks (\eg ``vest'' with 14 landmarks and ``skirt'' with 8 landmarks), the parameter space changes more significantly in those unseen tasks with more landmarks, which accounts for the increasing error on clothing categories with more landmarks.
MetaCloth achieves smaller error in the above cases, demonstrating its effectiveness in learning feature representations and adjusting the parameter space for unseen tasks.

\subsection{Ablation Study}
\subsubsection{Models}
During meta-training, after the parameters of the landmark detectors are predicted, MetaCloth optimizes the meta feature extraction network for initialization. During meta-test, the model is tuned with the meta-trained initialization from a few annotated samples in unseen tasks. 
To analyze the effects of the tuning scheme with the meta-trained initialization, we evaluate four different models in benchmark-2.

\paragraph{Base-FEN} We remove the optimization of the meta feature extraction network for initialization as Base-FEN. Specifically, this model only optimizes the parameters ${\theta}_0$ of the base feature extraction network and the parameters $\phi$ of the parameter prediction network following Alg.~\ref{alg:PPNet}, rather than train the meta feature extraction network with parameters $\theta$ for initialization. During meta-test, it is evaluated following Alg.~\ref{alg:test} except that in line 5, the parameters of the feature extraction network are ${\theta}_0$, which are trained on all seen categories with standard supervised learning.

\paragraph{LD-Keep} We keep the the optimization of the meta feature extraction network for initialization, but remove the update of the parameters for the landmark detectors during meta-training as LD-Keep. Specifically, this model is trained in the same way as MetaCloth following Alg.~\ref{alg:train} except that it does not update the parameters of the landmark detectors as line 7-9 before optimizing the parameters $\theta$. During meta-test, it is evaluated following Alg.~\ref{alg:test} as MetaCloth.

\paragraph{Base-FEN${\vartriangle}$ and LD-Keep${\vartriangle}$} We adopt the same training scheme as Base-FEN and LD-Keep, but remove the update of the parameters for the landmark detectors during meta-test as Base-FEN${\vartriangle}$ and LD-Keep${\vartriangle}$. Specifically, they do not update the parameters of the landmark detectors as line 6-8 in Alg.~\ref{alg:test} for unseen tasks.

\subsubsection{Results} Table~\ref{tab:abalation} presents the few-shot fashion landmark detection results in benchmark-2. We have the following observations.
First, during meta-test, the models Base-FEN and LD-Keep that update the parameters of the landmark detectors achieve better performance on all shots than the models Base-FEN${\vartriangle}$ and LD-Keep${\vartriangle}$ that do not. After the feature extraction network is tuned with a few annotated samples in a new task, it extracts task-specific landmark features. In this way, updating the parameters of the landmark detectors from those landmark features enables the landmark detectors to be task-specific as well. The improved performance demonstrates the effectiveness of the parameter prediction scheme, since the landmark detectors can be updated to generate better results after the feature extraction network is tuned.

\begin{table}\small\centering
	\vspace{-10pt}
	\caption{Ablation study on the effects of the tuning scheme with the meta-trained initialization in benchmark-2. The evaluation metric is normalized error (NE) with 95\% confidence intervals ($\times{10}^{-2}$), which is averaged over 700 episodes and \textbf{smaller value indicates better performance}. Compared with models from the first row to the fourth row, mean error on MetaCloth decreases by \textbf{13.9\%}, \textbf{17.7\%}, \textbf{9.7\%}, \textbf{21.8\%}.}
	\scalebox{0.8}{
		\begin{tabular}{c|ccccccc}
			\toprule[1.5pt]
			Benchmark-2&shot-3&shot-5&shot-8&shot-10&Mean\\
			\toprule[1.5pt]
			Base-FEN &0.137$\pm$2.5&0.113$\pm$2.3&0.094$\pm$1.9&0.088$\pm$1.8&0.108\\
			Base-FEN${\vartriangle}$&0.141$\pm$2.7&0.117$\pm$2.4&0.101$\pm$2.3&0.093$\pm$2.1&0.113\\
			LD-Keep&0.134$\pm$2.6&0.107$\pm$2.3&0.090$\pm$1.9&0.082$\pm$1.5&0.103\\
			LD-Keep${\vartriangle}$&0.148$\pm$2.8&0.124$\pm$2.7&0.105$\pm$2.3&0.100$\pm$2.2&0.119\\
			\toprule[1.5pt]
			MetaCloth&0.121$\pm$1.7&0.096$\pm$1.5&0.080$\pm$1.4&0.077$\pm$1.2&0.093\\
			\bottomrule[1.5pt]
	\end{tabular}}
	\vspace{-10pt}
	\label{tab:abalation}
\end{table}

Second, the model Base-FEN without the meta-trained initialization have larger error than MetaCloth. In MetaCloth, the parameters of the meta feature extraction network are explicitly trained on a large number of tasks for initialization, while Base-FEN only uses the parameters of the base feature extraction network for initialization, which is trained on all seen categories with standard supervised learning. The improved performance on MetaCloth over Base-FEN indicates that the model can be better adapted to unseen tasks from a few samples with the meta-trained initialization.

Finally, the model LD-Keep, which does not update the parameters of landmark detectors during meta-training, also degrades performance compared with MetaCloth. Since the parameters of the updated landmark detectors are predicted from landmark features extracted by the meta feature extraction network, the loss calculated using the parameters of the updated landmark detectors is actually only related to the parameters of the meta feature extraction network. In this way, MetaCloth optimizes the meta feature extraction network more effectively with the updated landmark detectors.

\begin{figure}[t]
	\vspace{-10pt}
	\begin{center}
		\includegraphics[width=0.8\linewidth]{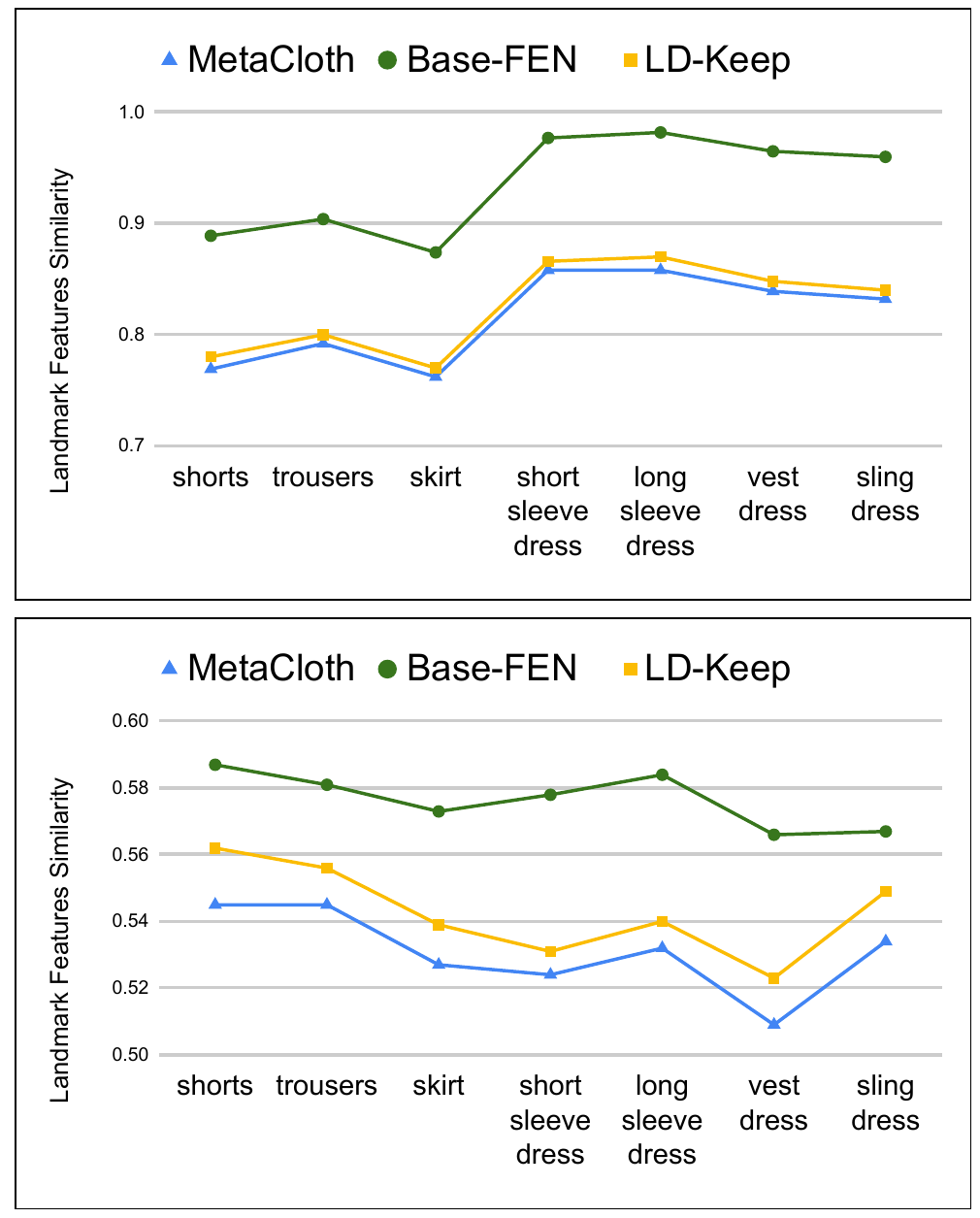}
	\end{center}
	\vspace{-10pt}
	\caption{The similarity between features of (a). the same landmark before and after tuning the model and (b). different landmarks from the tuned model on different unseen categories with 8 shots in benchmark-2. Smaller value in (a) indicates that the model is adjusted more significantly with a few annotated samples since the extracted landmark features change more after the model is tuned. Smaller value in (b) indicates that the model extracts more distinctive landmark features since features of different landmarks are better separated.}
	\vspace{-10pt}
	\label{fig:ablation}
\end{figure}

Following~\cite{raghu2019rapid,oh2020does}, we further show the similarity between features of the same landmarks before and after tuning the model, and of different landmarks from the tuned model on different unseen categories in Fig.~\ref{fig:ablation}. The experiments are conducted with 8 shots in benchmark-2. From Fig.~\ref{fig:ablation} (a), we can see that the similarity between features of the same landmarks before and after tuning the model Base-FEN is much larger, showing that the landmark features do not change a lot after the model is tuned. Without the meta-trained initialization, Base-FEN can not adjust the feature extraction network significantly to be adapted to a new task from a few annotated samples.
In Fig.~\ref{fig:ablation} (b), the features of different landmarks from the tuned MetaCloth have smaller similarity, and thus are better separated. Compared with LD-Keep, which do not update the parameters of landmark detectors to optimize the meta feature extraction network during meta-training, MetaCloth produces more effective optimization, such that the feature extraction network can be tuned from a few samples to extract more distinctive landmark features.

\section{Conclusion}
This work is among the first to explore few-shot dense fashion landmark detection with meta-learning in a $N_c$-way $K$-shot setup, where $N_c$ varies across tasks because various clothing categories have different number of landmarks. We propose an effective framework named MetaCloth, which can dynamically generate different numbers of parameters for different landmark detectors of unseen tasks, and learn a highly generalizable feature extraction network from only a few annotated samples with meta-learned initialization. Extensive evaluations are conducted on different benchmarks in DeepFashion2, showing the effectiveness of MetaCloth. Few-shot dense fashion landmark detection enables a learned model generalized to arbitrary clothing category with any number of landmarks by using only a few annotated images, and will facilitate the real-world applications to detect clothing landmarks without relying on a large number of annotations.

\small\noindent\textbf{IEEE Copyright} © 2021 IEEE.  Personal use of this material is permitted.  Permission from IEEE must be obtained for all other uses, in any current or future media, including reprinting/republishing this material for advertising or promotional purposes, creating new collective works, for resale or redistribution to servers or lists, or reuse of any copyrighted component of this work in other works.

\bibliographystyle{IEEEtran}
\bibliography{IEEEfull}

\begin{thebibliography}{10}
\providecommand{\url}[1]{#1}
\csname url@samestyle\endcsname
\providecommand{\newblock}{\relax}
\providecommand{\bibinfo}[2]{#2}
\providecommand{\BIBentrySTDinterwordspacing}{\spaceskip=0pt\relax}
\providecommand{\BIBentryALTinterwordstretchfactor}{4}
\providecommand{\BIBentryALTinterwordspacing}{\spaceskip=\fontdimen2\font plus
\BIBentryALTinterwordstretchfactor\fontdimen3\font minus
  \fontdimen4\font\relax}
\providecommand{\BIBforeignlanguage}[2]{{%
\expandafter\ifx\csname l@#1\endcsname\relax
\typeout{** WARNING: IEEEtran.bst: No hyphenation pattern has been}%
\typeout{** loaded for the language `#1'. Using the pattern for}%
\typeout{** the default language instead.}%
\else
\language=\csname l@#1\endcsname
\fi
#2}}
\providecommand{\BIBdecl}{\relax}
\BIBdecl

\bibitem{sun2019meta}
Q.~Sun, Y.~Liu, T.-S. Chua, and B.~Schiele, ``Meta-transfer learning for
  few-shot learning,'' in \emph{Proceedings of the IEEE Conference on Computer
  Vision and Pattern Recognition}, 2019, pp. 403--412.

\bibitem{wu2020meta}
X.~Wu, D.~Sahoo, and S.~Hoi, ``Meta-rcnn: Meta learning for few-shot object
  detection,'' in \emph{Proceedings of the 28th ACM International Conference on
  Multimedia}, 2020, pp. 1679--1687.

\bibitem{wang2019panet}
K.~Wang, J.~H. Liew, Y.~Zou, D.~Zhou, and J.~Feng, ``Panet: Few-shot image
  semantic segmentation with prototype alignment,'' in \emph{Proceedings of the
  IEEE International Conference on Computer Vision}, 2019, pp. 9197--9206.

\bibitem{liu2016deepfashion}
Z.~Liu, P.~Luo, S.~Qiu, X.~Wang, and X.~Tang, ``Deepfashion: Powering robust
  clothes recognition and retrieval with rich annotations,'' in
  \emph{Proceedings of the IEEE conference on computer vision and pattern
  recognition}, 2016, pp. 1096--1104.

\bibitem{ge2019deepfashion2}
Y.~Ge, R.~Zhang, X.~Wang, X.~Tang, and P.~Luo, ``Deepfashion2: A versatile
  benchmark for detection, pose estimation, segmentation and re-identification
  of clothing images,'' in \emph{Proceedings of the IEEE Conference on Computer
  Vision and Pattern Recognition}, 2019, pp. 5337--5345.

\bibitem{tryon}
D.~Roy, S.~Santra, and B.~Chanda, ``Lgvton: A landmark guided approach to
  virtual try-on,'' \emph{arXiv preprint arXiv:2004.00562}, 2020.

\bibitem{ge2021parser}
Y.~Ge, Y.~Song, R.~Zhang, C.~Ge, W.~Liu, and P.~Luo, ``Parser-free virtual
  try-on via distilling appearance flows,'' in \emph{Proceedings of the
  IEEE/CVF Conference on Computer Vision and Pattern Recognition}, 2021, pp.
  8485--8493.

\bibitem{ge2021disentangled}
C.~Ge, Y.~Song, Y.~Ge, H.~Yang, W.~Liu, and P.~Luo, ``Disentangled cycle
  consistency for highly-realistic virtual try-on,'' in \emph{Proceedings of
  the IEEE/CVF Conference on Computer Vision and Pattern Recognition}, 2021,
  pp. 16\,928--16\,937.

\bibitem{finn2017model}
C.~Finn, P.~Abbeel, and S.~Levine, ``Model-agnostic meta-learning for fast
  adaptation of deep networks,'' in \emph{Proceedings of the 34th International
  Conference on Machine Learning-Volume 70}.\hskip 1em plus 0.5em minus
  0.4em\relax JMLR. org, 2017, pp. 1126--1135.

\bibitem{nichol2018first}
A.~Nichol, J.~Achiam, and J.~Schulman, ``On first-order meta-learning
  algorithms,'' \emph{arXiv preprint arXiv:1803.02999}, 2018.

\bibitem{gidaris2019generating}
S.~Gidaris and N.~Komodakis, ``Generating classification weights with gnn
  denoising autoencoders for few-shot learning,'' \emph{arXiv preprint
  arXiv:1905.01102}, 2019.

\bibitem{baik2020meta}
S.~Baik, M.~Choi, J.~Choi, H.~Kim, and K.~M. Lee, ``Meta-learning with adaptive
  hyperparameters,'' \emph{arXiv preprint arXiv:2011.00209}, 2020.

\bibitem{kang2019few}
B.~Kang, Z.~Liu, X.~Wang, F.~Yu, J.~Feng, and T.~Darrell, ``Few-shot object
  detection via feature reweighting,'' in \emph{Proceedings of the IEEE
  International Conference on Computer Vision}, 2019, pp. 8420--8429.

\bibitem{wang2019meta}
Y.-X. Wang, D.~Ramanan, and M.~Hebert, ``Meta-learning to detect rare
  objects,'' in \emph{Proceedings of the IEEE International Conference on
  Computer Vision}, 2019, pp. 9925--9934.

\bibitem{fu2019meta}
K.~Fu, T.~Zhang, Y.~Zhang, M.~Yan, Z.~Chang, Z.~Zhang, and X.~Sun, ``Meta-ssd:
  Towards fast adaptation for few-shot object detection with meta-learning,''
  \emph{IEEE Access}, vol.~7, pp. 77\,597--77\,606, 2019.

\bibitem{dong2018few}
N.~Dong and E.~Xing, ``Few-shot semantic segmentation with prototype
  learning.'' in \emph{BMVC}, vol.~1, 2018, p.~6.

\bibitem{zhang2020sg}
X.~Zhang, Y.~Wei, Y.~Yang, and T.~S. Huang, ``Sg-one: Similarity guidance
  network for one-shot semantic segmentation,'' \emph{IEEE Transactions on
  Cybernetics}, 2020.

\bibitem{pambala2020sml}
A.~K. Pambala, T.~Dutta, and S.~Biswas, ``Sml: Semantic meta-learning for
  few-shot semantic segmentation,'' \emph{arXiv preprint arXiv:2009.06680},
  2020.

\bibitem{metadataset}
E.~Triantafillou, T.~Zhu, V.~Dumoulin, P.~Lamblin, U.~Evci, K.~Xu, R.~Goroshin,
  C.~Gelada, K.~Swersky, P.-A. Manzagol \emph{et~al.}, ``Meta-dataset: A
  dataset of datasets for learning to learn from few examples,'' \emph{arXiv
  preprint arXiv:1903.03096}, 2019.

\bibitem{toshev2014deeppose}
A.~Toshev and C.~Szegedy, ``Deeppose: Human pose estimation via deep neural
  networks,'' in \emph{Proceedings of the IEEE conference on computer vision
  and pattern recognition}, 2014, pp. 1653--1660.

\bibitem{cao2019openpose}
Z.~Cao, G.~Hidalgo, T.~Simon, S.-E. Wei, and Y.~Sheikh, ``Openpose: realtime
  multi-person 2d pose estimation using part affinity fields,'' \emph{IEEE
  transactions on pattern analysis and machine intelligence}, vol.~43, no.~1,
  pp. 172--186, 2019.

\bibitem{wang2018attentive}
W.~Wang, Y.~Xu, J.~Shen, and S.-C. Zhu, ``Attentive fashion grammar network for
  fashion landmark detection and clothing category classification,'' in
  \emph{Proceedings of the IEEE Conference on Computer Vision and Pattern
  Recognition}, 2018, pp. 4271--4280.

\bibitem{li2019spatial}
Y.~Li, S.~Tang, Y.~Ye, and J.~Ma, ``Spatial-aware non-local attention for
  fashion landmark detection,'' \emph{arXiv preprint arXiv:1903.04104}, 2019.

\bibitem{yu2019layout}
W.~Yu, X.~Liang, K.~Gong, C.~Jiang, N.~Xiao, and L.~Lin, ``Layout-graph
  reasoning for fashion landmark detection,'' in \emph{Proceedings of the IEEE
  Conference on Computer Vision and Pattern Recognition}, 2019, pp. 2937--2945.

\bibitem{kwitt2016one}
R.~Kwitt, S.~Hegenbart, and M.~Niethammer, ``One-shot learning of scene
  locations via feature trajectory transfer,'' in \emph{Proceedings of the IEEE
  Conference on Computer Vision and Pattern Recognition}, 2016, pp. 78--86.

\bibitem{hariharan2017low}
B.~Hariharan and R.~Girshick, ``Low-shot visual recognition by shrinking and
  hallucinating features,'' in \emph{Proceedings of the IEEE International
  Conference on Computer Vision}, 2017, pp. 3018--3027.

\bibitem{liu2018feature}
B.~Liu, X.~Wang, M.~Dixit, R.~Kwitt, and N.~Vasconcelos, ``Feature space
  transfer for data augmentation,'' in \emph{Proceedings of the IEEE conference
  on computer vision and pattern recognition}, 2018, pp. 9090--9098.

\bibitem{schwartz2018delta}
E.~Schwartz, L.~Karlinsky, J.~Shtok, S.~Harary, M.~Marder, A.~Kumar, R.~Feris,
  R.~Giryes, and A.~Bronstein, ``Delta-encoder: an effective sample synthesis
  method for few-shot object recognition,'' in \emph{Advances in Neural
  Information Processing Systems}, 2018, pp. 2845--2855.

\bibitem{pfister2014domain}
T.~Pfister, J.~Charles, and A.~Zisserman, ``Domain-adaptive discriminative
  one-shot learning of gestures,'' in \emph{European Conference on Computer
  Vision}.\hskip 1em plus 0.5em minus 0.4em\relax Springer, 2014, pp. 814--829.

\bibitem{wu2018exploit}
Y.~Wu, Y.~Lin, X.~Dong, Y.~Yan, W.~Ouyang, and Y.~Yang, ``Exploit the unknown
  gradually: One-shot video-based person re-identification by stepwise
  learning,'' in \emph{Proceedings of the IEEE Conference on Computer Vision
  and Pattern Recognition}, 2018, pp. 5177--5186.

\bibitem{douze2018low}
M.~Douze, A.~Szlam, B.~Hariharan, and H.~J{\'e}gou, ``Low-shot learning with
  large-scale diffusion,'' in \emph{Proceedings of the IEEE Conference on
  Computer Vision and Pattern Recognition}, 2018, pp. 3349--3358.

\bibitem{yan2015multi}
W.~Yan, J.~Yap, and G.~Mori, ``Multi-task transfer methods to improve one-shot
  learning for multimedia event detection.'' in \emph{BMVC}, 2015, pp. 37--1.

\bibitem{luo2017label}
Z.~Luo, Y.~Zou, J.~Hoffman, and L.~F. Fei-Fei, ``Label efficient learning of
  transferable representations acrosss domains and tasks,'' in \emph{Advances
  in neural information processing systems}, 2017, pp. 165--177.

\bibitem{motiian2017few}
S.~Motiian, Q.~Jones, S.~Iranmanesh, and G.~Doretto, ``Few-shot adversarial
  domain adaptation,'' in \emph{Advances in neural information processing
  systems}, 2017, pp. 6670--6680.

\bibitem{benaim2018one}
S.~Benaim and L.~Wolf, ``One-shot unsupervised cross domain translation,'' in
  \emph{Advances in Neural Information Processing Systems}, 2018, pp.
  2104--2114.

\bibitem{fink2004object}
M.~Fink, ``Object classification from a single example utilizing class
  relevance metrics,'' \emph{Advances in neural information processing
  systems}, vol.~17, pp. 449--456, 2004.

\bibitem{koch2015siamese}
G.~Koch, R.~Zemel, and R.~Salakhutdinov, ``Siamese neural networks for one-shot
  image recognition,'' in \emph{ICML deep learning workshop}, vol.~2.\hskip 1em
  plus 0.5em minus 0.4em\relax Lille, 2015.

\bibitem{triantafillou2017few}
E.~Triantafillou, R.~Zemel, and R.~Urtasun, ``Few-shot learning through an
  information retrieval lens,'' in \emph{Advances in Neural Information
  Processing Systems}, 2017, pp. 2255--2265.

\bibitem{oreshkin2018tadam}
B.~Oreshkin, P.~R. L{\'o}pez, and A.~Lacoste, ``Tadam: Task dependent adaptive
  metric for improved few-shot learning,'' in \emph{Advances in Neural
  Information Processing Systems}, 2018, pp. 721--731.

\bibitem{xu2017few}
Z.~Xu, L.~Zhu, and Y.~Yang, ``Few-shot object recognition from machine-labeled
  web images,'' in \emph{Proceedings of the IEEE Conference on Computer Vision
  and Pattern Recognition}, 2017, pp. 1164--1172.

\bibitem{zhu2018compound}
L.~Zhu and Y.~Yang, ``Compound memory networks for few-shot video
  classification,'' in \emph{Proceedings of the European Conference on Computer
  Vision (ECCV)}, 2018, pp. 751--766.

\bibitem{cai2018memory}
Q.~Cai, Y.~Pan, T.~Yao, C.~Yan, and T.~Mei, ``Memory matching networks for
  one-shot image recognition,'' in \emph{Proceedings of the IEEE conference on
  computer vision and pattern recognition}, 2018, pp. 4080--4088.

\bibitem{ramalho2019adaptive}
T.~Ramalho and M.~Garnelo, ``Adaptive posterior learning: few-shot learning
  with a surprise-based memory module,'' \emph{arXiv preprint
  arXiv:1902.02527}, 2019.

\bibitem{salakhutdinov2012one}
R.~Salakhutdinov, J.~Tenenbaum, and A.~Torralba, ``One-shot learning with a
  hierarchical nonparametric bayesian model,'' in \emph{Proceedings of ICML
  Workshop on Unsupervised and Transfer Learning}, 2012, pp. 195--206.

\bibitem{rezende2016one}
D.~J. Rezende, S.~Mohamed, I.~Danihelka, K.~Gregor, and D.~Wierstra, ``One-shot
  generalization in deep generative models,'' \emph{arXiv preprint
  arXiv:1603.05106}, 2016.

\bibitem{reed2017few}
S.~Reed, Y.~Chen, T.~Paine, A.~v.~d. Oord, S.~Eslami, D.~Rezende, O.~Vinyals,
  and N.~de~Freitas, ``Few-shot autoregressive density estimation: Towards
  learning to learn distributions,'' \emph{arXiv preprint arXiv:1710.10304},
  2017.

\bibitem{zhang2018metagan}
R.~Zhang, T.~Che, Z.~Ghahramani, Y.~Bengio, and Y.~Song, ``Metagan: An
  adversarial approach to few-shot learning,'' \emph{Advances in Neural
  Information Processing Systems}, vol.~31, pp. 2365--2374, 2018.

\bibitem{shaban2017one}
A.~Shaban, S.~Bansal, Z.~Liu, I.~Essa, and B.~Boots, ``One-shot learning for
  semantic segmentation,'' \emph{arXiv preprint arXiv:1709.03410}, 2017.

\bibitem{li2017meta}
Z.~Li, F.~Zhou, F.~Chen, and H.~Li, ``Meta-sgd: Learning to learn quickly for
  few-shot learning,'' \emph{arXiv preprint arXiv:1707.09835}, 2017.

\bibitem{rajeswaran2019meta}
A.~Rajeswaran, C.~Finn, S.~M. Kakade, and S.~Levine, ``Meta-learning with
  implicit gradients,'' in \emph{Advances in Neural Information Processing
  Systems}, 2019, pp. 113--124.

\bibitem{banerjee2020meta}
A.~Banerjee, ``Meta-drn: Meta-learning for 1-shot image segmentation,''
  \emph{arXiv preprint arXiv:2008.00247}, 2020.

\bibitem{vinyals2016matching}
O.~Vinyals, C.~Blundell, T.~Lillicrap, D.~Wierstra \emph{et~al.}, ``Matching
  networks for one shot learning,'' in \emph{Advances in neural information
  processing systems}, 2016, pp. 3630--3638.

\bibitem{snell2017prototypical}
J.~Snell, K.~Swersky, and R.~Zemel, ``Prototypical networks for few-shot
  learning,'' in \emph{Advances in Neural Information Processing Systems},
  2017, pp. 4077--4087.

\bibitem{sung2018learning}
F.~Sung, Y.~Yang, L.~Zhang, T.~Xiang, P.~H. Torr, and T.~M. Hospedales,
  ``Learning to compare: Relation network for few-shot learning,'' in
  \emph{Proceedings of the IEEE Conference on Computer Vision and Pattern
  Recognition}, 2018, pp. 1199--1208.

\bibitem{bertinetto2016learning}
L.~Bertinetto, J.~F. Henriques, J.~Valmadre, P.~Torr, and A.~Vedaldi,
  ``Learning feed-forward one-shot learners,'' \emph{Advances in neural
  information processing systems}, vol.~29, pp. 523--531, 2016.

\bibitem{gidaris2018dynamic}
S.~Gidaris and N.~Komodakis, ``Dynamic few-shot visual learning without
  forgetting,'' in \emph{Proceedings of the IEEE Conference on Computer Vision
  and Pattern Recognition}, 2018, pp. 4367--4375.

\bibitem{hypernetworks}
D.~Ha, A.~Dai, and Q.~V. Le, ``Hypernetworks,'' \emph{arXiv preprint
  arXiv:1609.09106}, 2016.

\bibitem{tail}
Y.-X. Wang, D.~Ramanan, and M.~Hebert, ``Learning to model the tail,'' in
  \emph{Proceedings of the 31st International Conference on Neural Information
  Processing Systems}, 2017, pp. 7032--7042.

\bibitem{analogy}
L.~Zhou, P.~Cui, S.~Yang, W.~Zhu, and Q.~Tian, ``Learning to learn image
  classifiers with visual analogy,'' in \emph{Proceedings of the IEEE/CVF
  Conference on Computer Vision and Pattern Recognition}, 2019, pp.
  11\,497--11\,506.

\bibitem{motion}
L.-Y. Gui, Y.-X. Wang, D.~Ramanan, and J.~M. Moura, ``Few-shot human motion
  prediction via meta-learning,'' in \emph{Proceedings of the European
  Conference on Computer Vision (ECCV)}, 2018, pp. 432--450.

\bibitem{regression}
Y.-X. Wang and M.~Hebert, ``Learning to learn: Model regression networks for
  easy small sample learning,'' in \emph{European Conference on Computer
  Vision}.\hskip 1em plus 0.5em minus 0.4em\relax Springer, 2016, pp. 616--634.

\bibitem{qiao2018few}
S.~Qiao, C.~Liu, W.~Shen, and A.~L. Yuille, ``Few-shot image recognition by
  predicting parameters from activations,'' in \emph{Proceedings of the IEEE
  Conference on Computer Vision and Pattern Recognition}, 2018, pp. 7229--7238.

\bibitem{he2016deep}
K.~He, X.~Zhang, S.~Ren, and J.~Sun, ``Deep residual learning for image
  recognition,'' in \emph{Proceedings of the IEEE conference on computer vision
  and pattern recognition}, 2016, pp. 770--778.

\bibitem{deng2009imagenet}
J.~Deng, W.~Dong, R.~Socher, L.-J. Li, K.~Li, and L.~Fei-Fei, ``Imagenet: A
  large-scale hierarchical image database,'' in \emph{2009 IEEE conference on
  computer vision and pattern recognition}.\hskip 1em plus 0.5em minus
  0.4em\relax Ieee, 2009, pp. 248--255.

\bibitem{he2017real}
Y.~He, L.~Yang, and L.~Chen, ``Real-time fashion-guided clothing semantic
  parsing: a lightweight multi-scale inception neural network and benchmark,''
  in \emph{Workshops at the Thirty-First AAAI Conference on Artificial
  Intelligence}, 2017.

\bibitem{raghu2019rapid}
A.~Raghu, M.~Raghu, S.~Bengio, and O.~Vinyals, ``Rapid learning or feature
  reuse? towards understanding the effectiveness of maml,'' \emph{arXiv
  preprint arXiv:1909.09157}, 2019.

\bibitem{oh2020does}
J.~Oh, H.~Yoo, C.~Kim, and S.-Y. Yun, ``Does maml really want feature reuse
  only?'' \emph{arXiv preprint arXiv:2008.08882}, 2020.

\end{thebibliography}

\end{document}